\documentclass[sigconf]{aamas} 

\renewcommand*\backref[1]{\ifx#1\relax \else (Cited in Section #1) \fi}
\usepackage{url}
\usepackage{graphicx}
\usepackage{float}
\usepackage{xcolor}
\usepackage{xkcdcolors} 
\usepackage{algorithm}
\usepackage{algcompatible}
\usepackage{amsmath, amssymb, amsfonts, amsthm}
\usepackage{balance} 
\usepackage{enumerate}

\definecolor{relevantgreen}{RGB}{0, 204, 0}  
\definecolor{noisered}{RGB}{204, 0, 0}   

\newcommand\sbullet[1][.5]{\mathbin{\vcenter{\hbox{\scalebox{#1}{$\bullet$}}}}}

\setcopyright{ifaamas}
\acmConference[AAMAS '23]{Proc.\@ of the 22nd International Conference
on Autonomous Agents and Multiagent Systems (AAMAS 2023)}{May 29 -- June 2, 2023}
{London, United Kingdom}{A.~Ricci, W.~Yeoh, N.~Agmon, B.~An (eds.)}
\copyrightyear{2023}
\acmYear{2023}
\acmDOI{}
\acmPrice{}
\acmISBN{}

\acmSubmissionID{432}

\title[Automatic Noise Filtering]{Automatic Noise Filtering with Dynamic Sparse Training\\in Deep Reinforcement Learning}

\author{Bram Grooten}
\affiliation{
  \institution{Eindhoven University of Technology}
  }
\email{b.j.grooten@tue.nl}

\author{Ghada Sokar}
\affiliation{
  \institution{Eindhoven University of Technology}
}
\email{g.a.z.n.sokar@tue.nl}

\author{Shibhansh Dohare}
\affiliation{
  \institution{University of Alberta}
}
\email{dohare@ualberta.ca}

\author{Elena Mocanu}
\affiliation{
  \institution{University of Twente}
}
\email{e.mocanu@utwente.nl}

\author{Matthew E. Taylor}
\affiliation{
  \institution{University of Alberta \& Alberta Machine Intelligence Institute (Amii)}
}
\email{matthew.e.taylor@ualberta.ca}

\author{Mykola Pechenizkiy}
\affiliation{
  \institution{Eindhoven University of Technology}
}
\email{m.pechenizkiy@tue.nl}

\author{Decebal Constantin Mocanu}
\affiliation{
  \institution{University of Luxembourg \& University of Twente}
  \city{}
  \country{}}
\email{d.c.mocanu@utwente.nl}

\begin{abstract}
Tomorrow's robots will need to distinguish useful information from noise when performing different tasks. A household robot for instance may continuously receive a plethora of information about the home, but needs to focus on just a small subset to successfully execute its current chore.
Filtering distracting inputs that contain irrelevant data has received little attention in the reinforcement learning literature. 
To start resolving this, we formulate a problem setting in reinforcement learning called the \textit{extremely noisy environment} (ENE), where up to 99\% of the input features are pure noise.
Agents need to detect which features provide task-relevant information about the state of the environment. 
Consequently, we propose a new method termed \textit{Automatic Noise Filtering} (ANF), which uses the principles of dynamic sparse training in synergy with various deep reinforcement learning algorithms. The sparse input layer learns to focus its connectivity on task-relevant features, such that ANF-SAC and ANF-TD3 outperform standard SAC and TD3 by a large margin, while using up to 95\% fewer weights.
Furthermore, we devise a transfer learning setting for ENEs, by permuting all features of the environment after 1M timesteps to simulate the fact that other information sources can become relevant as the world evolves. Again, ANF surpasses the baselines in final performance and sample complexity. Our code is available online.\footnote{See \url{https://github.com/bramgrooten/automatic-noise-filtering}}
\end{abstract}


\newcommand{\BibTeX}{\rm B\kern-.05em{\sc i\kern-.025em b}\kern-.08em\TeX}

\begin{document}

\pagestyle{fancy}
\fancyhead{}

\maketitle

\thispagestyle{plain}  
\pagestyle{plain}

\section{Introduction}
\label{sec:intro}

Future robots will likely perceive a plethora of information about the state of the world, but only parts of it are going to be relevant to their current task.
For instance, a household robot receiving abundant information about all objects and processes in the house.\footnote{For example: cleanliness of floors, furniture, cupboards, kitchen utensils; CO$_2$, CO levels and temperature in each room; up-to-date stock of all food and non-food items in the fridge and/or basement; mood, nourishment, and health of all inhabitants; etc.} 
For its current task, e.g. making pancakes, only a small subset of these information sources, or \textit{features}, are relevant. Agents should automatically detect which features are task-relevant, without humans having to predefine this. Other examples may be: a hearing aid distinguishing between voices and auditory noise, a surgical robot receiving all possible information about the patient, or a self-driving car that needs to ignore distracting billboards.

\begin{figure}
    \centering
    \includegraphics[width=0.9\linewidth]{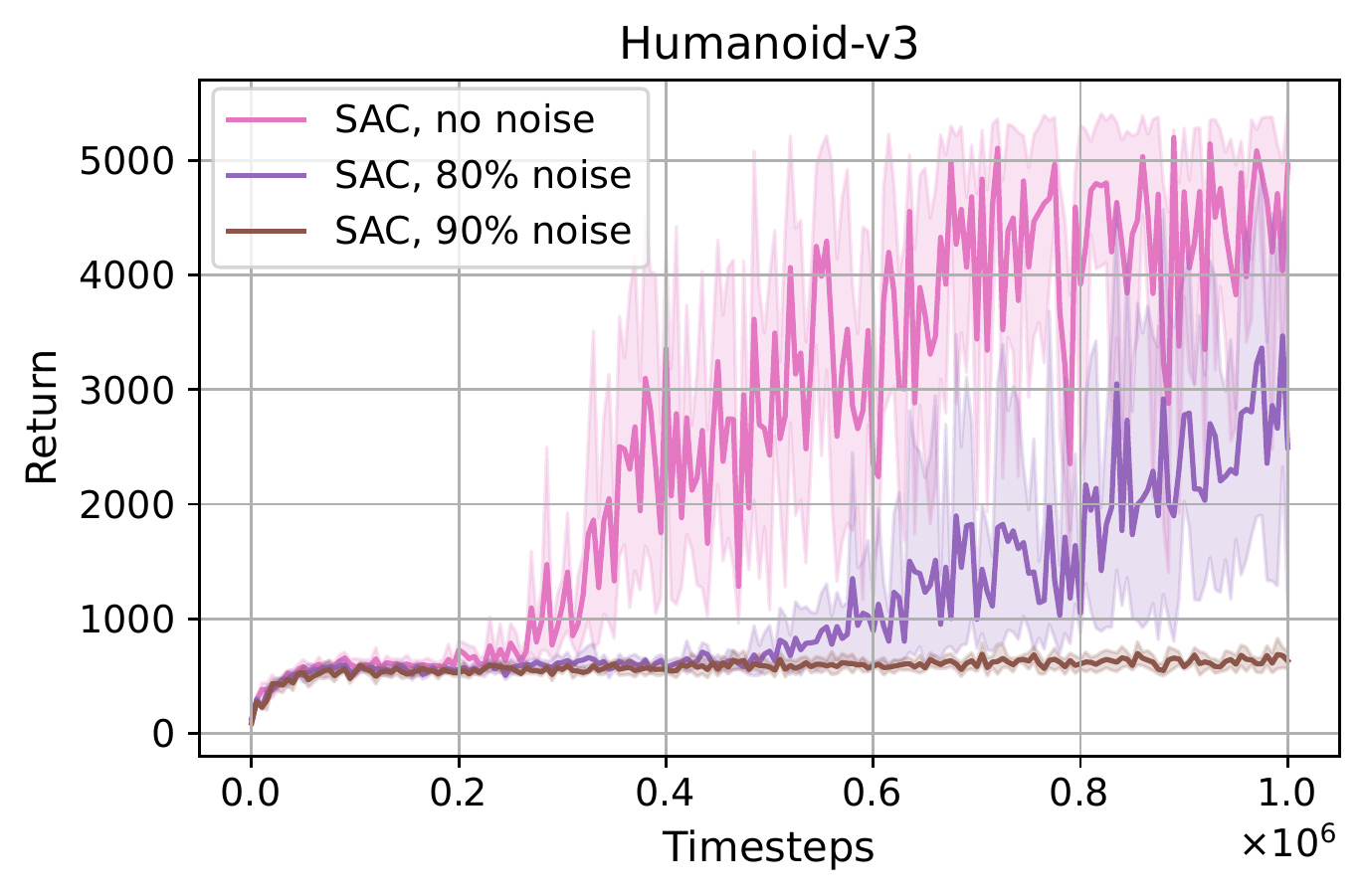}
    \vspace{-0.5cm}
    \caption{Performance of SAC on Humanoid-v3 environments expanded with a different number of pure noise features.
    Once the environment contains too much noise, SAC struggles to learn a decent policy. Standard dense networks cannot filter through the noise well enough on this problem.} 
    \vspace{-0.35cm}
    \label{fig:SACnoise}
\end{figure}

To illustrate the current situation: Soft Actor-Critic (SAC) \cite{sac} fails to learn a decent policy on an environment with 90\% added noise features, see \mbox{Figure~\ref{fig:SACnoise}}.
We simulate the noisy real-world environment by adding synthetic noise features to an existing state space. This allows us to study the problem in a controlled environment to understand where we stand and what can be done.
We need to invent methods that can effectively filter through the noise while learning to perform the environment's task.
Our \mbox{\textbf{research~question}} becomes: \textit{How can we design RL agents to learn and perform well in an extremely noisy environment?}

Dynamic Sparse Training (DST), a class of methods stemming from the Sparse Evolutionary Training (SET) algorithm \cite{set}, is promising in this regard. 
By starting from a randomly sparsified network and subsequently pruning and growing connections (weights) during training, DST searches for the optimal network topology.
DST is able to perform efficient feature selection for unsupervised learning, as shown by \citep{quickselection, sokar2022where}.
Further, \cite{mintaskrepr} discovered that sparse networks can find minimal task representations in deep RL by pruning redundant input dimensions. 
Not long after, \cite{ghada} successfully applied DST in deep RL, reducing the number of parameters without compromising performance.

This leads us to a plausible approach to our research question.
We think that the adaptability of DST can improve an agent’s sparse network structure such that task-relevant features are emphasized by receiving more connections than noise features. The combination of sparsity and adaptability enables the agent to filter through the noise more effectively, outperforming dense network approaches. 
The underlying hypothesis follows:

\vspace{0.5em}
\textbf{The Adaptability Hypothesis}:
\textit{A sparse neural network layer can adapt the location of its connections (weights) to gain a better performance \underline{faster} than a dense layer can adapt the weight values to achieve the same gain.}
\vspace{0.5em}

Note that newly grown connections still need to adjust their weight values through gradient descent, but we hypothesize that this generally happens quicker than a dense network modifies all of its weights. Relocated weights may receive a more informative gradient when connected to task-relevant features.
Briefly, the hypothesis states: dropping and growing connections is easier than adjusting the weights. This is inspired by our own brain's plasticity, which also dynamically drops and grows synapses \cite{neuroplasticity60s, castaldi2020neuroplasticity, pereda2014electrical}.

To verify our hypothesis, we propose a new algorithm called Automatic Noise Filtering (ANF), which can easily be combined with deep RL methods. It has a sparse input layer with adapting connectivity through dynamic sparse training. We compare ANF to two strong baseline deep RL algorithms: SAC \cite{sac} and TD3 \cite{td3}, which have fully dense layers throughout their networks.
We devise the \textit{extremely noisy environment} (ENE), further defined in Section~\ref{sec:problem}, which expands the state space of an existing RL environment with a large number of noise features. We apply this approach to four continuous control tasks from MuJoCo Gym \cite{mujoco, gym}. 


\paragraph{\textbf{Contributions.}}
\begin{itemize}
    \item We formulate a problem setting termed the \textit{extremely noisy environment} (ENE), where up to 99\% of the input features consist of pure noise. Agents need to detect the task-relevant features autonomously.
    \item We propose Automatic Noise Filtering (ANF), a dynamic sparse training method that outperforms baseline deep RL algorithms by a large margin, especially on environments with high noise levels.
    \item We devise a transfer learning setting of extremely noisy environments and show that ANF has better performance and forward transfer than the baselines SAC and TD3.
    \item We show that highly sparse ANF agents with up to 95\% fewer parameters can still surpass their dense baselines on the extremely noisy environments.
    \item We extend the ENE by adjusting the noise distribution in two ways, increasing the difficulty. ANF maintains its advantage on these challenging extensions.\footnote{See an illustrative video here: \url{https://youtu.be/vS47UnsTQk8}}
\end{itemize}

\paragraph{\textbf{Outline.}}
In Section~\ref{sec:problem} we formulate the problem setting. Section~\ref{sec:rel-work} gives an overview of the background and related work. Our method is introduced in Section~\ref{sec:anf}, along with the first experiments. In Section~\ref{sec:transfer} we explore the transfer learning setting. Sections~\ref{sec:louder} through \ref{sec:sparse} provide further analysis, where we perform an ablation study and discover how far we can extend our problem and algorithm. Finally, Section~\ref{sec:conclusion} concludes the paper. Additional results, details, and discussion are in the Appendix.%


\section{Problem Formulation}
\label{sec:problem}

We introduce a problem setting where agents have to act in environments that contain a lot of noise. As the noise features generally greatly outnumber the task-relevant features in this setting, we simply call it the extremely noisy environment (ENE).

\begin{figure}
    \centering
    \includegraphics[width=\linewidth]{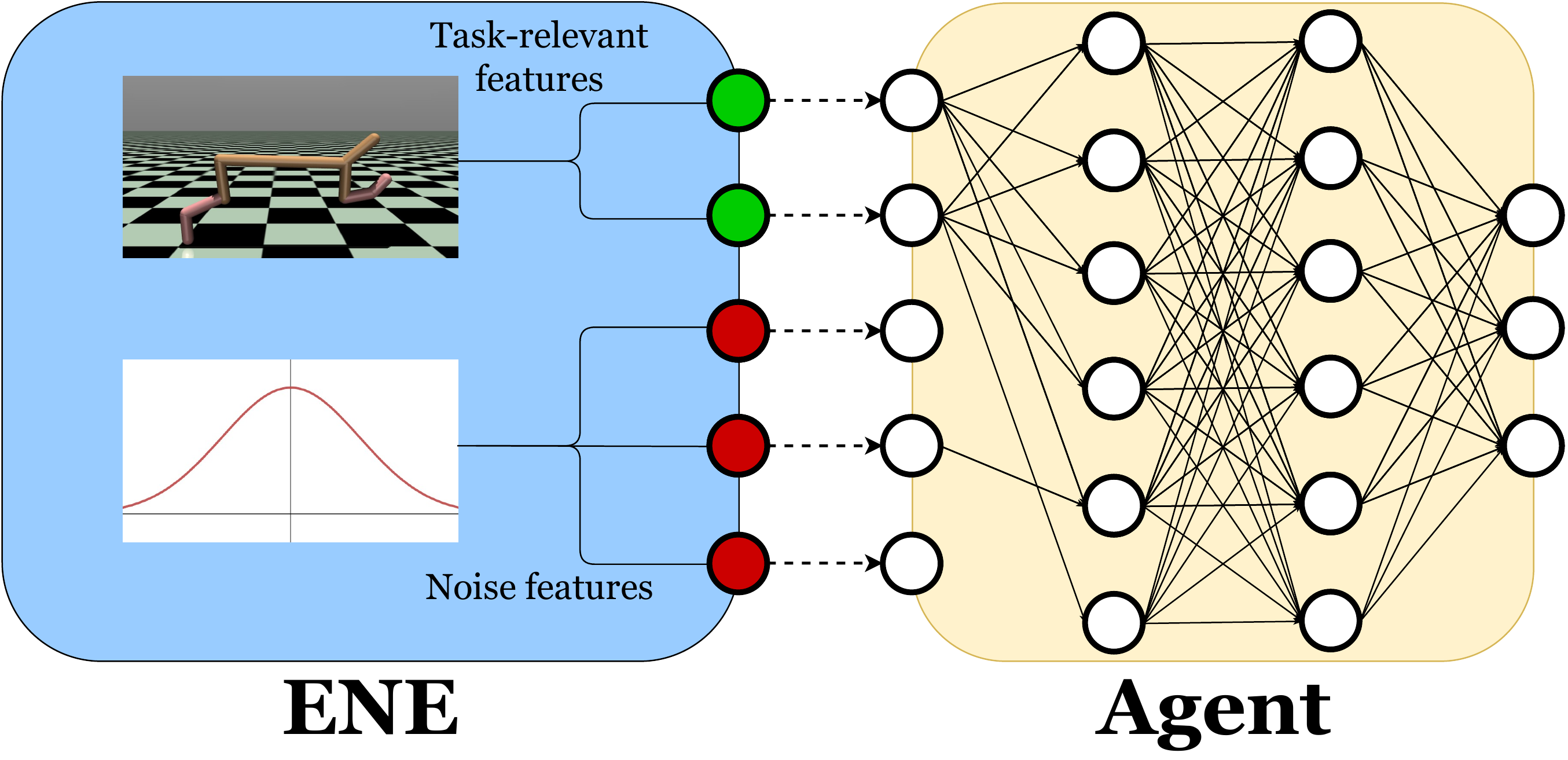}
    \vspace{-0.5cm}
    \caption{Extremely noisy environments (ENEs) contain many noise 
    {\color{noisered}$\sbullet[1.8]$}  
    features and some relevant 
    {\color{relevantgreen}$\sbullet[1.8]$}  
    features. We use ENEs where up to $99\%$ of the features are noise. Our method Automatic Noise Filtering (ANF) learns to predominantly connect with the input neurons that provide useful information and outperforms dense baselines by a large margin, especially in the noisiest environments.}
    \label{fig:noisefeats}
\end{figure}

\textbf{\textit{Extremely noisy environment.}}
To create an ENE, we take any reinforcement learning environment that generates feature vectors as states. The ENE expands this feature vector by concatenating many additional features consisting only of pure noise, sampled from any given distribution. 
An agent is not told which features are useful (task-relevant) and which are useless (noise), so it has to learn to ignore the distracting noise features by itself, see Figure~\ref{fig:noisefeats}.


\begin{figure*}
    \centering
    \includegraphics[width=\textwidth]{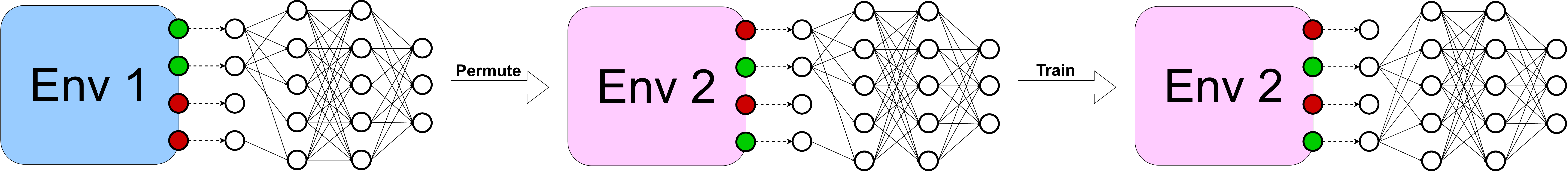}
    \caption{In \textit{permutated} extremely noisy environments (PENE) the order of the input features gets shuffled after a certain number of timesteps. Our ANF agents automatically adjust their network structure to this new environment. They show superior forward transfer compared to fully dense methods, even though ANF agents might need to prune and regrow many connections in the sparse input layer. We hypothesize that adapting the location of sparse connections is easier than adjusting the weights of all connections in a dense network.}
    \label{fig:pene}
\end{figure*}

In our main experiments, the noise features produce pure Gaussian noise, sampled i.i.d. from $\mathcal{N}(0,1)$. The fraction of noise features in an ENE is denoted by $n_f \in [0,1)$. For example, for $n_f = 0.5$ we enlarge the original state space of a MuJoCo Gym environment by lengthening the state feature vector by a factor of 2. 
In general, the dimensionality of the new state space is
\begin{equation*}
    d_{ene} = \left\lceil \dfrac{d_{og}}{1- n_f} \right\rceil 
\end{equation*}
where $d_{og}$ is the number of dimensions in the original state space. As $n_f$ increases to $1$, the dimensionality of the ENE expands.

\textbf{\textit{Transfer learning setting.}}
Next to the ENE, we introduce an even more challenging problem setting where, after every $T_p$ timesteps, all input features are permuted at random. This permutation simulates the fact that other features can become relevant over time. 
Previously irrelevant features might suddenly become relevant, for \mbox{example}, when a household robot gets a new task.\footnote{Features could also gradually become more relevant, as the world evolves (i.e. concept drift). This is outside the scope of our research, we focus on the sudden change.}
In our case, the change in environment is \textit{not} announced to the agent. Agents need to detect the change and transfer their representations quickly to adapt to the new instantiation of the \textit{permutated extremely noisy environment} (PENE), see Figure~\ref{fig:pene}.
A previously task-relevant feature may or may not still be relevant after the permutation, inducing the need to rediscover the distribution of the features and filter through the noise.


\section{Background and Related Work}
\label{sec:rel-work}

Our proposed ANF algorithm is based on dynamic sparse training (DST). In this section, we briefly overview the related work of DST in reinforcement learning (RL) and existing noise filtering methods.

\textbf{\textit{Sparse training.}}
Dynamic sparse training is a subfield of the sparse training regime \cite{mocanu_aamas}, where weights deemed superfluous are pruned away to increase the efficiency of a neural network. In dense-to-sparse training, dense networks are gradually pruned to higher sparsity levels throughout training \cite{han, lth, dst-but-dense-to-sparse}. In sparse-to-sparse training, where DST belongs, a network begins with a high sparsity level from scratch \cite{set, deeprewiring}. The existing connections can either stay fixed (\textit{static} sparse training) or be pruned and regrown during training (\textit{dynamic} sparse training).

In supervised learning, especially computer vision, many promising results have been achieved with sparsity over the last few years \cite{itop, rigl, vit_sparse}. These algorithms benefit from potential performance boosts, decreased computational costs, and better generalization \citep{generalize, generalization}. Furthermore, DST has been used successfully for an efficient feature selection algorithm \cite{quickselection}, which inspired our project.

\textbf{\textit{DST in RL.}}
Applying sparse training in reinforcement learning is useful, as real-world applications often deal with latency constraints \cite{plasma}, which limits the number of parameters.
Unfortunately, in the area of RL it seems that applying sparse training is more challenging than in supervised learning, as the achievable sparsity levels without loss in performance are generally lower \cite{state-of-sparseRL}. 
Only a few papers have applied sparse training to deep RL so far. In the \textit{offline} RL setting, \cite{offlineRL} have reached 95\% sparsity with almost no performance degradation. While this is impressive, we believe that offline RL is more similar to supervised learning than online RL. Moreover, it does not support learning in changing environments \citep{reward_is_enough}. Therefore, we focus on the \textit{online} RL setting throughout the paper, and even go into the transfer learning setting \cite{taylor2009transfer}.

To the best of our knowledge, the first work applying DST to online RL is from \cite{ghada}. They outperform dense networks with the algorithms DS-TD3 and DS-SAC, which combine sparse evolutionary training (SET) \citep{set} with TD3 \citep{td3} and SAC \citep{sac}. The methods of \cite{ghada} form the foundation of our ANF algorithm. 

Sokar et al. \cite{ghada} reached a global sparsity level of 50\%, which was later improved upon by \cite{state-of-sparseRL, rlx2}, who experimented with sparsity levels up to 99\%. They showed that the sparsity level reachable without loss of performance largely depends on the environment.
Graesser et al. \cite{state-of-sparseRL} compared DST methods such as SET \cite{set} and RigL \cite{rigl} in many deep RL environments. Their performance proved to be quite similar, so we choose to use only SET.
\textbf{\textit{Noise in RL.} }
There exist different types of noise that an agent may encounter. Let us characterize the two main categories:
\begin{itemize}
    \item Type 1: uncertainty in perception, for example when an automated vehicle cannot clearly see a traffic sign since the sun is right next to it. 
    \item Type 2: distracting, task-irrelevant percepts, for example the bright colors of a billboard when crossing Times Square in New York City. 
\end{itemize}

Type 1 noise, i.e. measurement errors, is often researched by adding noise \textit{on top of} existing features to produce more robust agents \citep{wouter_robust, vinitsky2020robust, sun2021exploring, robust_survey}. This type of noise is outside the scope of this work.
Instead, we focus on type 2 noise and investigate it by adding synthetic features \textit{alongside} the existing features, creating a state space of higher dimensionality. The goal is to discover algorithms that can perform tasks well while having access to all available features, without having to pre-select the task-relevant ones. Feature selection should be carried out automatically by the RL agents.

\begin{figure*}
    \centering
    \includegraphics[width=\textwidth]{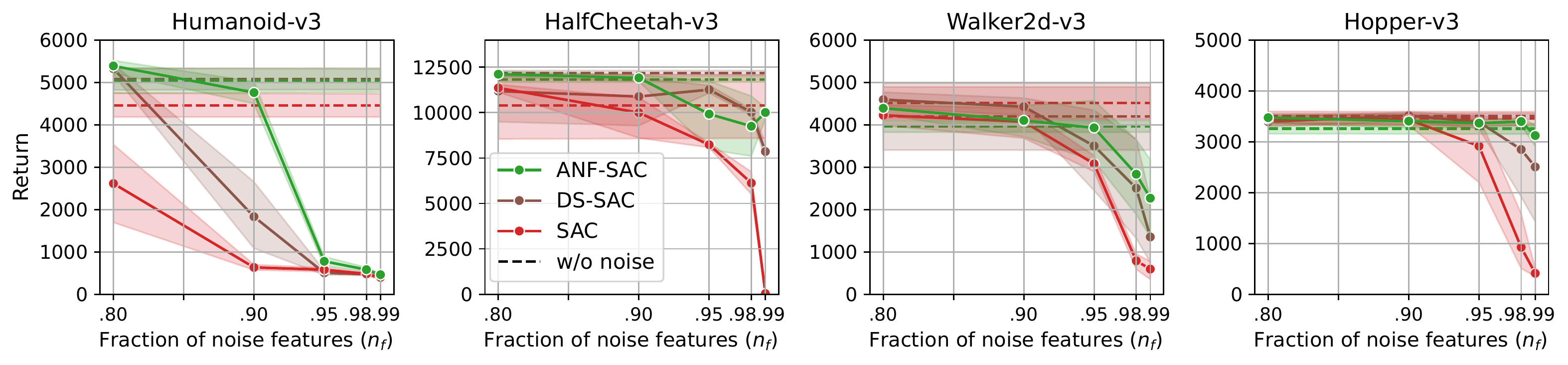}
    \vspace{-0.55cm}
    \caption{Performance of \textit{Automatic Noise Filtering} (ANF) compared to its baselines for different noise fractions $n_f$. 
    The curves show average return in the last 10\% of training over 5 seeds, with shaded regions representing 95\% confidence intervals.
    When the number of noise features in an environment increases, the performance of the standard fully-dense networks of SAC deteriorates much faster than ANF-SAC.
    Similar graphs for ANF-TD3 are shown in Figure~\ref{fig:overview80-99_horiline_TD3} of Appendix~\ref{sec:app-anf}.}
    \label{fig:overview80-99_horiline_SAC}
\end{figure*}

To the best of our knowledge, the first work to make an existing RL environment noisier by adding extra features was FS-NEAT \cite{fs-neat}. It introduced an evolutionary algorithm to select relevant features. Most of the follow-up work takes this evolutionary approach \cite{kroon2009automatic, bishop_evoluationary, acunto2012feature}, while we use the efficiency of deep learning, stochastic gradient descent, and dynamic sparse training.

Our work extends environments that provide the current state as a feature vector. However, it is worth mentioning that environments with visual (pixel) inputs have likewise been augmented to include a noisy challenge, such as distracting backgrounds \cite{stone2021distracting}.
Other methods that have some similarities to our approach include recognizing distractor signals \cite{rafiee2020eye_distractor}, reducing state dimensionality \cite{curran2016dimensionality, botteghi2021low}, and identifying fake features in federated learning \cite{li2021privacy}.




\section{Automatic Noise Filtering}
\label{sec:anf}

In this section, we explain how our ANF algorithm works, after which we show and interpret the results of our main experiments. ANF is a simple method that can be applied to any MLP-based deep RL algorithm. It is built upon the DS-TD3 and DS-SAC algorithms of \cite{ghada}, which use sparse evolutionary training (SET) from \cite{set} as the underlying dynamic sparse training method.

In both the actor and critic networks, ANF begins by randomly pruning the input layer to the desired sparsity level $s_i$.
During training, we drop weak connections of the input layer (weights with the smallest magnitude) after every topology-change period $\Delta T$.
After dropping a certain fraction $d_f$ of the existing weights, ANF randomly grows the same number of connections to maintain the sparsity level $s_i$.
By giving new connections enough time to increase their weights, ANF detects task-relevant features without explicit supervision. We provide pseudocode for ANF-SAC in Appendix~\ref{sec:app-pseudocode}. 

One aspect that sets ANF apart from the previous works on non-noisy settings \cite{ghada, state-of-sparseRL} is that we only sparsify the input layer. This helps us to pinpoint the support of DST on our Adaptability Hypothesis.
Furthermore, in extremely noisy environments it is essential to filter through the large fraction of noise. Dynamic sparse training can perform this filtering elegantly. It works well to focus the DST principle on the first layer only, as this is where the distinction between relevant and noise features is made. In Section~\ref{sec:sparse} we investigate models that also have sparse hidden layers. 

Another difference between ANF and DS-TD3/SAC \cite{ghada} is that we mask the running averages of first and second raw moments of the gradient within the Adam optimizer \cite{adam} for non-existing connections. When connections are dropped and later regrown, they do not have access to previous information if implemented in a truly sparse manner. 
This aspect has been overlooked in the implementation of some sparsity research papers that apply Adam and only simulate true sparsity with binary masks on top of the weight matrices. Our research also utilizes such binary masks while keeping the truly sparse implementation in mind. See Appendix~\ref{sec:app-hardware} for further discussion.

\textbf{\textit{Experimental setup.}}
We integrate our ANF method in two popular deep RL algorithms: SAC and TD3. This means we compare the algorithms ANF-SAC and ANF-TD3 with their fully-dense counterparts as baselines. Furthermore, we compare to the closely related DS-SAC and DS-TD3, which both use their default global sparsity level of $50\%$. 
All neural networks have two hidden layers of 256 neurons with the ReLU activation function.
After a hyperparameter search for ANF, we set the input layer sparsity $s_i$ to $80\%$, the topology-change period $\Delta T = 1000$ timesteps, and the drop fraction $d_f = 0.05$. 
Further hyperparameter settings replicate prior work \cite{ghada, sac, td3}. See Appendix~\ref{sec:app-hyper} for additional details.

Our experiments are carried out in four continuous control environments from the  MuJoCo Gym suite: Humanoid-v3, HalfCheetah-v3, Walker2d-v3, and Hopper-v3. We first run an experiment without any added noise features as a baseline and then start increasing the noise level. 
The fraction of noise features, $n_f$, ranges over the set $\{0.8, 0.9, 0.95, 0.98, 0.99\}$. Note that the state spaces of these settings increase by $5\times, 10\times, 20\times, 50\times$, and $100\times$, respectively.

We train our agents for 1 million timesteps and evaluate them by running 10 test episodes after every 5000 timesteps.
We measure the average return over the last 10\% of training, as done in \cite{state-of-sparseRL}, for
overview graphs such as Figure~\ref{fig:overview80-99_horiline_SAC}.
Throughout the paper, we run 5 random seeds for every setting. In the graphs, we show the average curve as well as a 95\% confidence interval. 

\begin{figure*}
    \centering
    \includegraphics[width=0.9\textwidth]{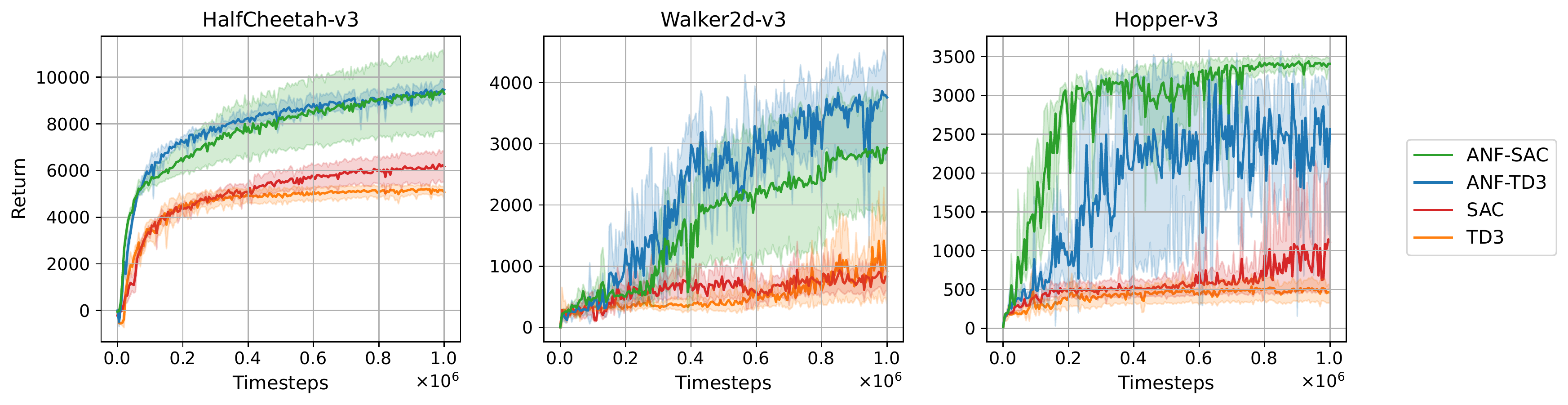}
    \vspace{-0.3cm}
    \caption{Learning curves on 3 environments with \underline{98\%} noise features. 
    This level of noise is too high for standard SAC and TD3 to learn a decent policy. ANF can still filter through the distraction and find a well-performing policy.}
    \vspace{-0.1cm}
    \label{fig:3envs-curves-fake98}
\end{figure*}

\begin{table}[b]
\caption{State and action space dimensions.}
\label{tab:dims}
\vspace{-0.2cm}
\setlength{\tabcolsep}{4pt}
\resizebox{\columnwidth}{!}{%
\begin{tabular}{lllll}\toprule
Environment    & State dim. & Action dim. & State dim. & State dim. \\
               & \small{\textit{Original}} & \small{\textit{Original}} & \small{\textit{ENE ($n_f\!=\!.8$)}}   & \small{\textit{ENE ($n_f\!=\!.99$)}}  \\  \midrule
Humanoid-v3    & 376        & 17          & 1880     & 37600        \\ 
HalfCheetah-v3 & 17         & 6           & 85       & 1700        \\ 
Walker2d-v3    & 17         & 6           & 85       & 1700        \\ 
Hopper-v3      & 11         & 3           & 55       & 1100        \\ \bottomrule         
\end{tabular}%
}
\end{table}

\textbf{\textit{Results.}}
First of all, the horizontal lines in Figure~\ref{fig:overview80-99_horiline_SAC} show that even in environments without noise ANF-SAC is able to reach similar or better performance than SAC and DS-SAC for \mbox{Humanoid-v3} and \mbox{HalfCheetah-v3}. By adjusting the connectivity of the input layer, ANF is able to select the set of most important features, the so-called \textit{minimal task representation} \cite{mintaskrepr}. 

Furthermore, when the noise level increases our ANF method outperforms the dense baseline by a significant margin on all environments. Especially in the noisiest environments, when $n_f \geq .95$, a large gap is visible between ANF-SAC and SAC for HalfCheetah, Walker2d, and Hopper. The Humanoid environment is an exception, as ANF outperforms its baseline much earlier here but then struggles with the high noise levels as well. Table~\ref{tab:dims} shows that Humanoid-v3 differs noticeably from the other three environments by the size of its state space.

\begin{figure}
    \centering
    \includegraphics[width=0.9\linewidth]{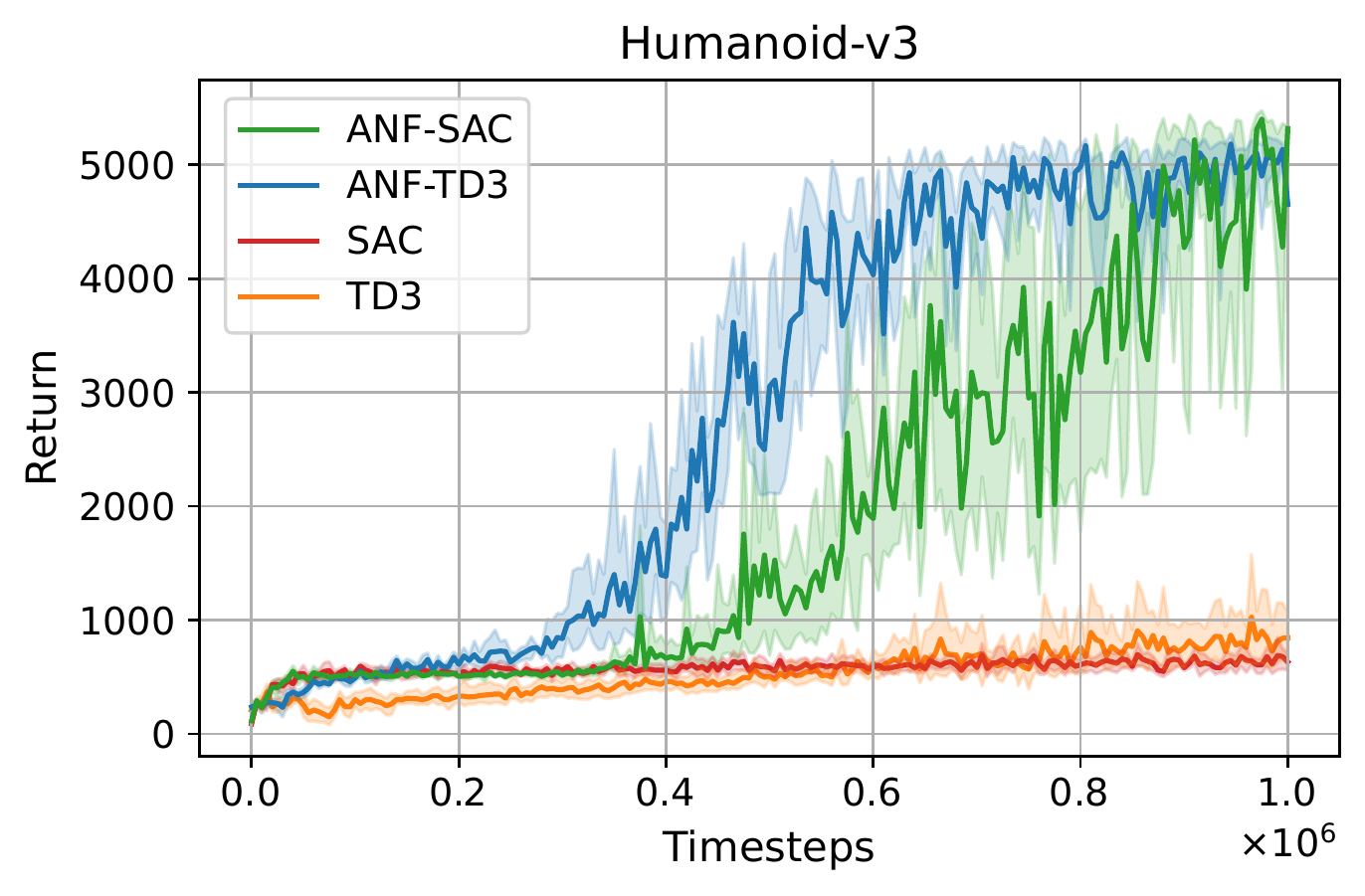}
    \vspace{-0.4cm}
    \caption{Learning curves for Humanoid-v3 with \underline{90\%} noise features. While standard SAC and TD3 are too distracted by the noise to learn the task, ANF finds the task-relevant features and is able to improve.}
    \vspace{-0.4cm}
    \label{fig:Humanoid-curves-fake90}
\end{figure}

The learning curves in Figure~\ref{fig:Humanoid-curves-fake90} indicate that SAC and TD3 are unable to learn a decent policy within 1M timesteps in this challenging extremely noisy environment. ANF learns to ignore the distracting noise and reaches a performance level \textit{similar even to SAC and TD3 in the environment without noise}.%
\footnote{Which is a return of $\sim$4500, see SAC's dashed line in Figure~\ref{fig:overview80-99_horiline_SAC}, Humanoid-v3.}
In the environments HalfCheetah, Walker2d, and Hopper, we continue to observe this behavior up to a noise fraction of 98\%, as shown in Figure~\ref{fig:3envs-curves-fake98}. ANF outperforms its baselines by a large margin in each environment.

\begin{figure}
    \centering
    \includegraphics[width=0.95\linewidth]{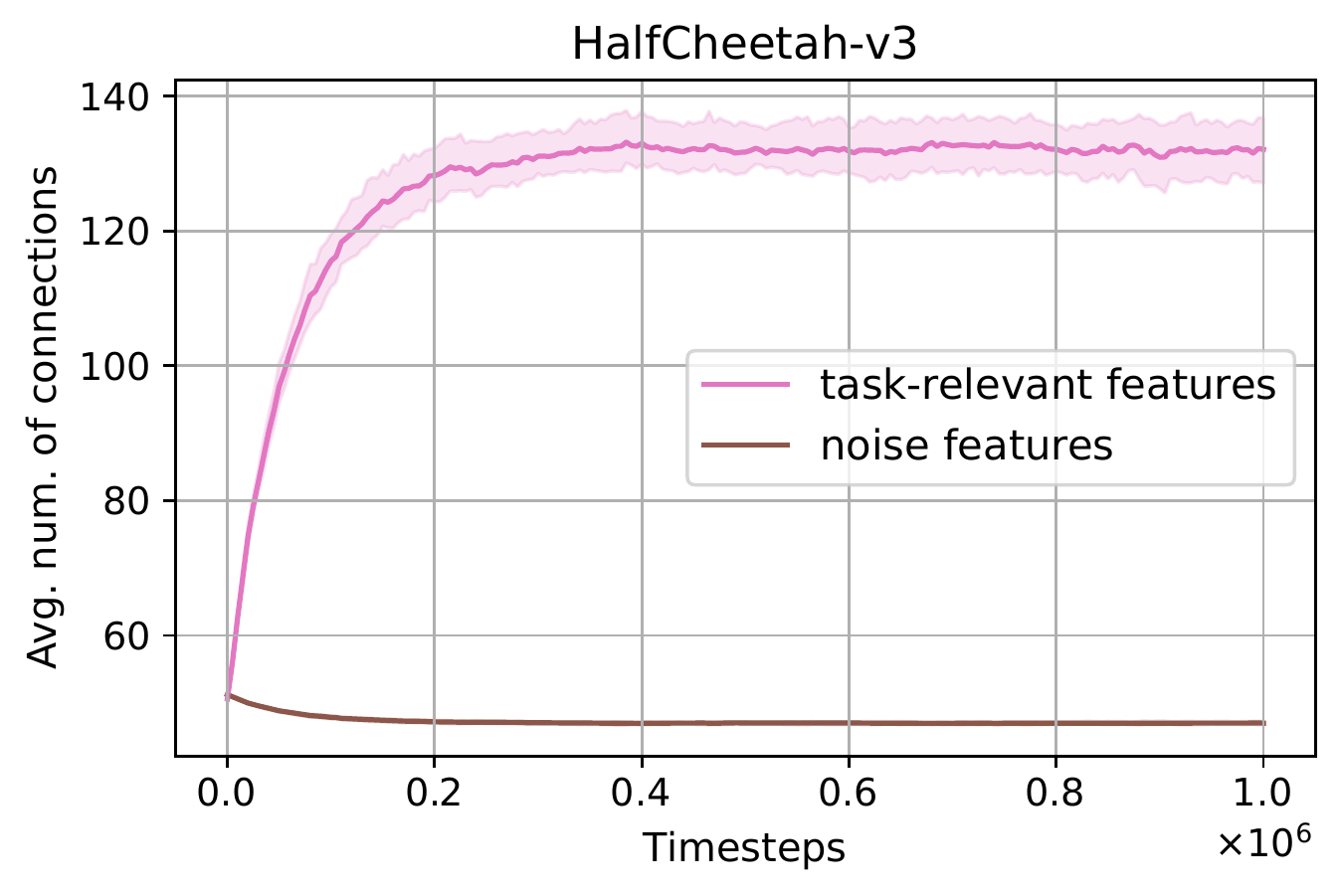}
    \vspace{-0.4cm}
    \caption{Average number of connections in the input layer of one of ANF-TD3's critic networks, on HalfCheetah-v3 with 90\% noise features. At the start of training every input neuron has around $256 \cdot 0.2 \approx 51$ connections, because the input layer sparsity is 80\% and connections are allocated uniformly at random. During training, ANF gradually prunes connections from the noise features and grows connections to the relevant features. A similar graph for ANF-SAC is shown in Figure~\ref{fig:connectivity_over_time_HalfCheetah_SAC_critic1} of Appendix~\ref{sec:app-anf}.}
    \label{fig:connectivity_over_time_HalfCheetah_TD3_critic1}
    \vspace{-0.4cm}
\end{figure}

\textbf{\textit{Topology shift.}}
To analyze what is actually happening, we visualize the connectivity of ANF. In Figure~\ref{fig:connectivity_over_time_HalfCheetah_TD3_critic1}, we present a graph that shows the development of the network's topology over time. 
The graph clearly demonstrates a topology shift in the input layer: 
on the one hand,
the average number of connections to task-relevant features rises, while
on the other hand,
noise features receive fewer weights. Together with the increased performance shown in Figure~\ref{fig:overview80-99_horiline_SAC}, this fully supports our Adaptivity Hypothesis.

\begin{figure*}
    \centering
    \includegraphics[width=0.87\textwidth]{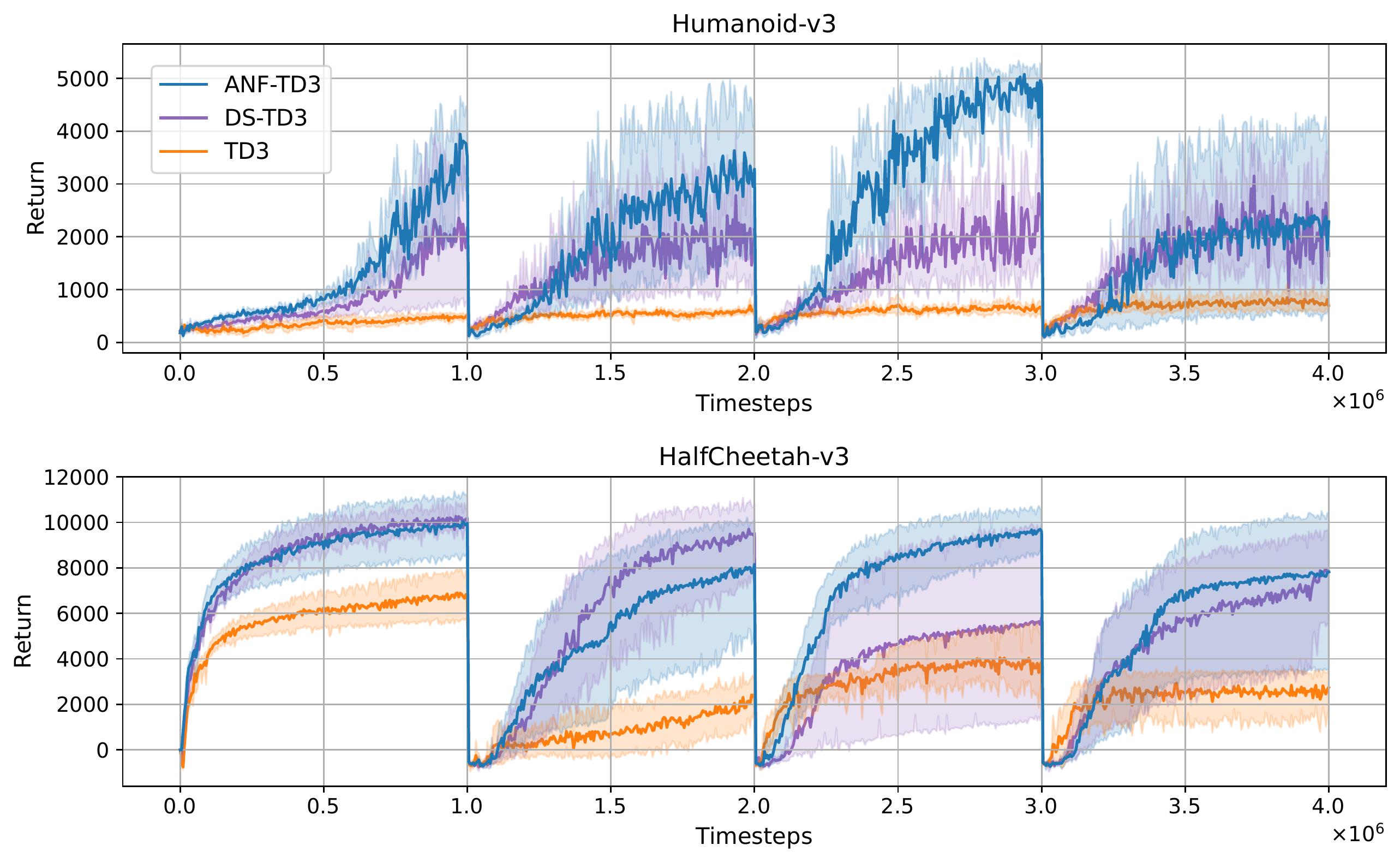}
    \vspace{-0.4cm}
    \caption{Performance of ANF-TD3 and its baselines on permuted extremely noisy environments (PENE) with 95\% noise features. After every 1M timesteps, the environment's features are shuffled with a random permutation. ANF is able to cope with this challenge, while the fully dense networks of TD3 are struggling. DS-TD3 performs decently, but ANF has an advantage by focusing its sparsity on the input layer. Similar graphs for ANF-SAC and other environments are shown in Appendix~\ref{sec:app-transfer}.}
    \vspace{-0.1cm}
    \label{fig:transfer_HumanoidHalfCheetah_TD3}
\end{figure*}

\section{Transfer Learning}
\label{sec:transfer}

During a robot's lifetime, it may happen that other information sources become relevant to its task. Moreover, the agent may receive an entirely new task, which can require it to focus its attention on totally different state features. 

We simulate this change in a \textit{permutated extremely noisy environment} (PENE), as described in Section~\ref{sec:problem}. The PENE rearranges all input features with a fixed permutation after every $T_p$ timesteps.
For our experiments, this means that the relevant and noise features are now mixed instead of concatenated. The agents will have to rediscover which input neurons are receiving task-relevant signals. Note that the PENE setting does \textit{not} announce the change in environment to the agent. 

\textbf{\textit{Experimental setup.}}
We set $T_p$ to 1M timesteps, such that agents have enough time to learn.
We run on the same four environments with a noise fraction of $n_f = 0.95$. In these experiments, we now train for 4 million timesteps, meaning that agents encounter four different instances (sub-environments) of feature permutations. 
Similar to the experiments of Section~\ref{sec:anf}, we compare ANF-SAC and ANF-TD3 with their fully dense baselines and DS-SAC/TD3. We show 95\% confidence intervals over 5 seeds. 


\textbf{\textit{Results.}}
Figure~\ref{fig:transfer_HumanoidHalfCheetah_TD3} shows the results for ANF-TD3 on Humanoid and HalfCheetah. See Appendix~\ref{sec:app-transfer} for the graphs of the remaining algorithms and environments.
It is evident that the performance drops considerably after each permutation of features. However, ANF is able to recover faster than the dense baselines in all environments. The method does not need to be adjusted for the challenging PENE setting; ANF keeps adapting the sparse input layer as before.

For Humanoid, some beneficial internal representations may be transferred forward, as the performance increases much earlier in the third sub-environment (between 2M and 3M timesteps) than when it is trained from scratch (between 0 and 1M timesteps). However, on the fourth sub-environment some ANF agents struggled a bit: each random seed determines not only the initialization of the agent, but also the random permutations of the environment. Thus, some sub-environments can be more challenging than others.

\textbf{\textit{Maintaining plasticity.}}
Agents that have to learn continually must be able to maintain plasticity. Standard methods are unable to do so, as shown by \cite{cbp}.
Since the connections of the input layer can drop and grow dynamically, ANF ensures that the agent has sufficient adaptability to adjust to a new environment. 
We analyze this plasticity by looking into the connectivity of the input layer once more, as done earlier in Figure~\ref{fig:connectivity_over_time_HalfCheetah_TD3_critic1} for the ENE experiments.

\begin{figure}
    \centering
    \includegraphics[width=.96\linewidth]{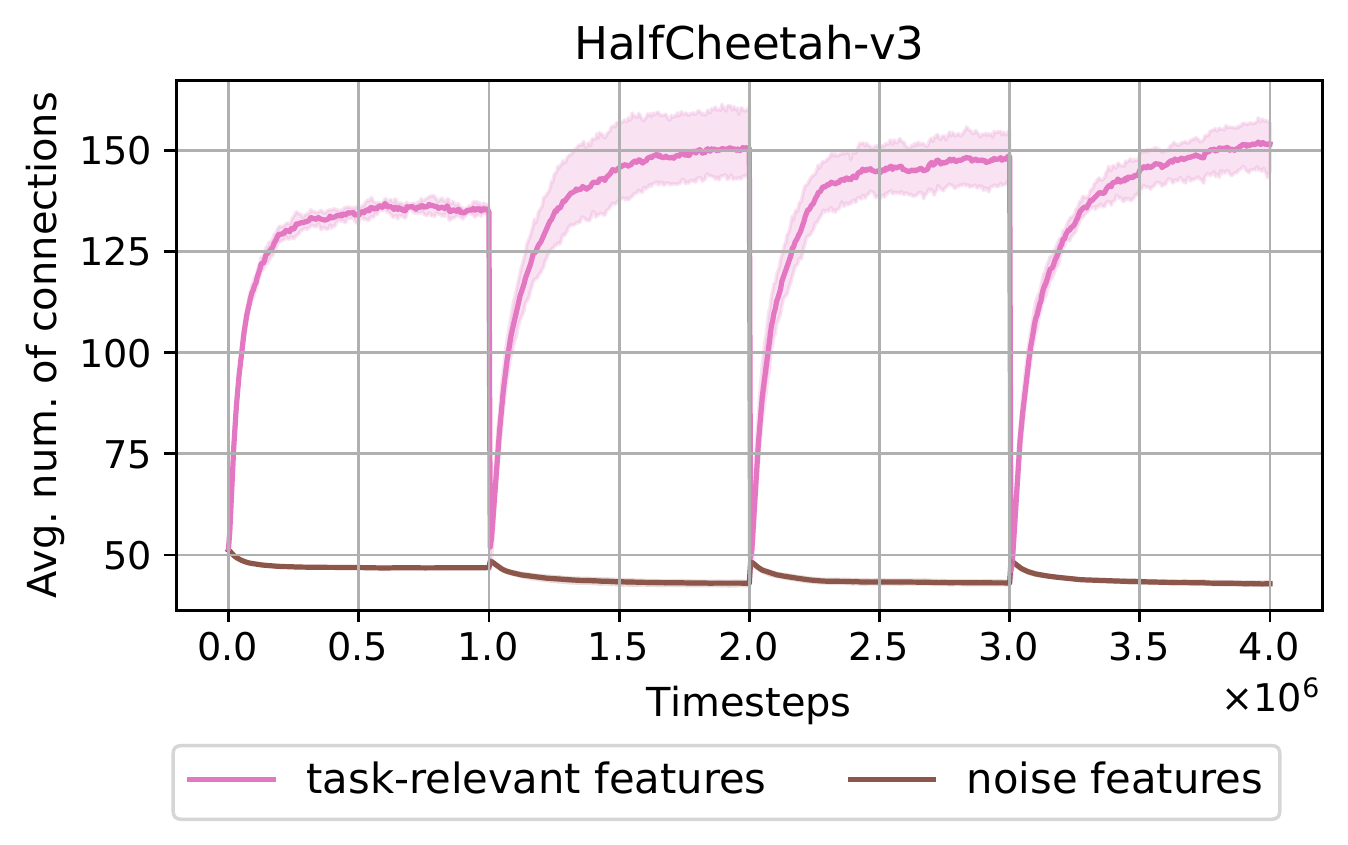}
    \vspace{-0.3cm}
    \caption{Average number of connections in the input layer of an ANF-SAC critic network, on HalfCheetah-v3 with $n_f = 0.95$. 
    Every 1M timesteps, the PENE permutes the order of the features. The ANF agent adjusts its network structure quickly, growing connections to the task-relevant features, which are now sent to different input neurons. 
    A similar graph for ANF-TD3 is shown in Figure~\ref{fig:connectivity_over_4Mtime_HalfCheetah_TD3_critic1} of Appendix~\ref{sec:app-transfer}.}
    \vspace{-0.2cm}
    \label{fig:connectivity_over_4Mtime_HalfCheetah_SAC_critic1}
\end{figure}

Now, in Figure~\ref{fig:connectivity_over_4Mtime_HalfCheetah_SAC_critic1}, we see that the average number of connections to task-relevant features quickly recovers after an environment change in the PENE. At every 1M steps, the PENE shuffles the features, which makes the average number of connections to relevant features drop considerably, close to the initial value. This is because many task-relevant signals are now coming in at input neurons that were previously receiving noise.

The fact that ANF has not pruned all connections to the irrelevant noise features after training on the first sub-environment is actually an advantage in this PENE setting. It means that ANF may be able to reach a high number of connections to new task-relevant features faster, as they already have some `spare' connections waiting.


\begin{figure}
    \centering
    \vspace{-0.1cm}
    \includegraphics[width=0.9\linewidth]{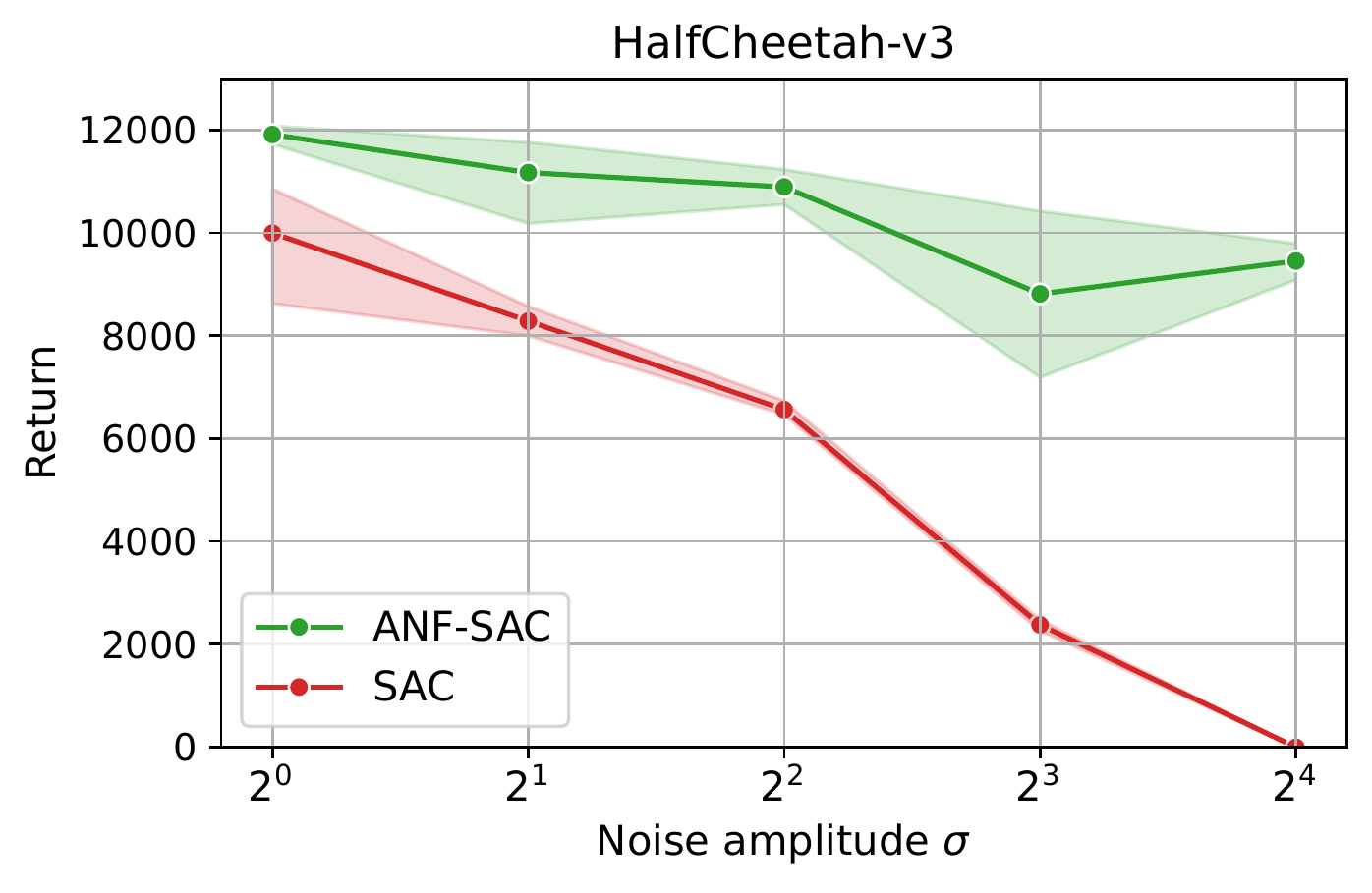}
    \vspace{-0.3cm}
    \caption{ANF-SAC and its dense baseline on ENEs with louder noise. The noise features are sampled from $\mathcal{N}(0, \sigma^2)$. Noise amplitude $\sigma$ is increased exponentially, notice the log-scale on the horizontal axis. 
    ANF tolerates louder noise much better than standard SAC, maintaining a high performance up to $\sigma = 16$. In this experiment $n_f\!=\!0.9$.
    The graph for ANF-TD3 is given in Figure \ref{fig:louder1_16_TD3} of Appendix~\ref{sec:app-louder}.}
    \vspace{-0.3cm}
    \label{fig:louder1_16_SAC}
\end{figure}

\section{Louder noise}
\label{sec:louder}

All of our experiments so far have been executed with noise features sampled from the standard Gaussian distribution of $\mathcal{N}(0,1)$. But what would happen if we increase the standard deviation, i.e. the noise amplitude? We expect the louder noise to be more distracting, 
increasing the difficulty of the ENE. 
We hope to discover whether ANF can cope with this additional challenge.

\textbf{\textit{Experimental setup.}}
We run ANF and its dense baselines on the ENE of HalfCheetah-v3, but the noise is now sampled from $\mathcal{N}(0,\sigma^2)$. We let the standard deviation $\sigma$ increase exponentially, ranging over the set $\{1, 2, 4, 8, 16\}$. The ENE contains 90\% of these louder noise features ($n_f = 0.9$).

\textbf{\textit{Results.}}
From the experimental results, we can conclude that louder noise does make the ENE more challenging. In Figure~\ref{fig:louder1_16_SAC}, it is clearly visible that as the noise amplitude increases, the performance decreases. Fortunately for ANF, this decrease is much less pronounced compared to its dense baseline. In fact, the final return of ANF-SAC on an ENE with noise amplitude $\sigma = 16$ is almost the same as SAC's performance on the standard $\sigma=1$ environment. 
ANF can cope well with even the loudest noise.\footnote{See \url{https://youtu.be/vS47UnsTQk8} for a video comparing the actual motion of \mbox{HalfCheetah} when controlled by ANF-SAC vs. SAC with different noise amplitudes.}

\section{Imitating real features}
\label{sec:imitate}

Upon closer inspection of the data distribution of the original state features, we discovered that these are far from a standard Gaussian distribution. In Appendix~\ref{sec:app-distr}, we present visualizations of the distributions of task-relevant features before and after training.\footnote{The challenging distribution shift of RL is clearly visible!}
To get closer to real-world noise, we want the noise features of our ENEs to imitate the original features. 
We expect that this increases the difficulty of our extremely noisy environments, as the noise is now much more similar to the task-relevant features.

\textbf{\textit{Experimental setup.}}
For each of the original features, we make a histogram of its final distribution (after training an agent in the standard environment), as shown in Appendix~\ref{sec:app-distr}. In the experiments of this section, the ENE samples from these histograms\footnote{By first sampling a bin according to the histogram's probability mass function, and then sampling a value uniformly at random within the chosen bin.} to generate noise features for the next state. We repeatedly sample from the distribution of each original feature until we have enough noise features. 
We run this experiment on HalfCheetah-v3, with 90\% of these noise features that imitate the task-relevant features. 

\begin{figure}
    \centering
    \includegraphics[width=0.9\linewidth]{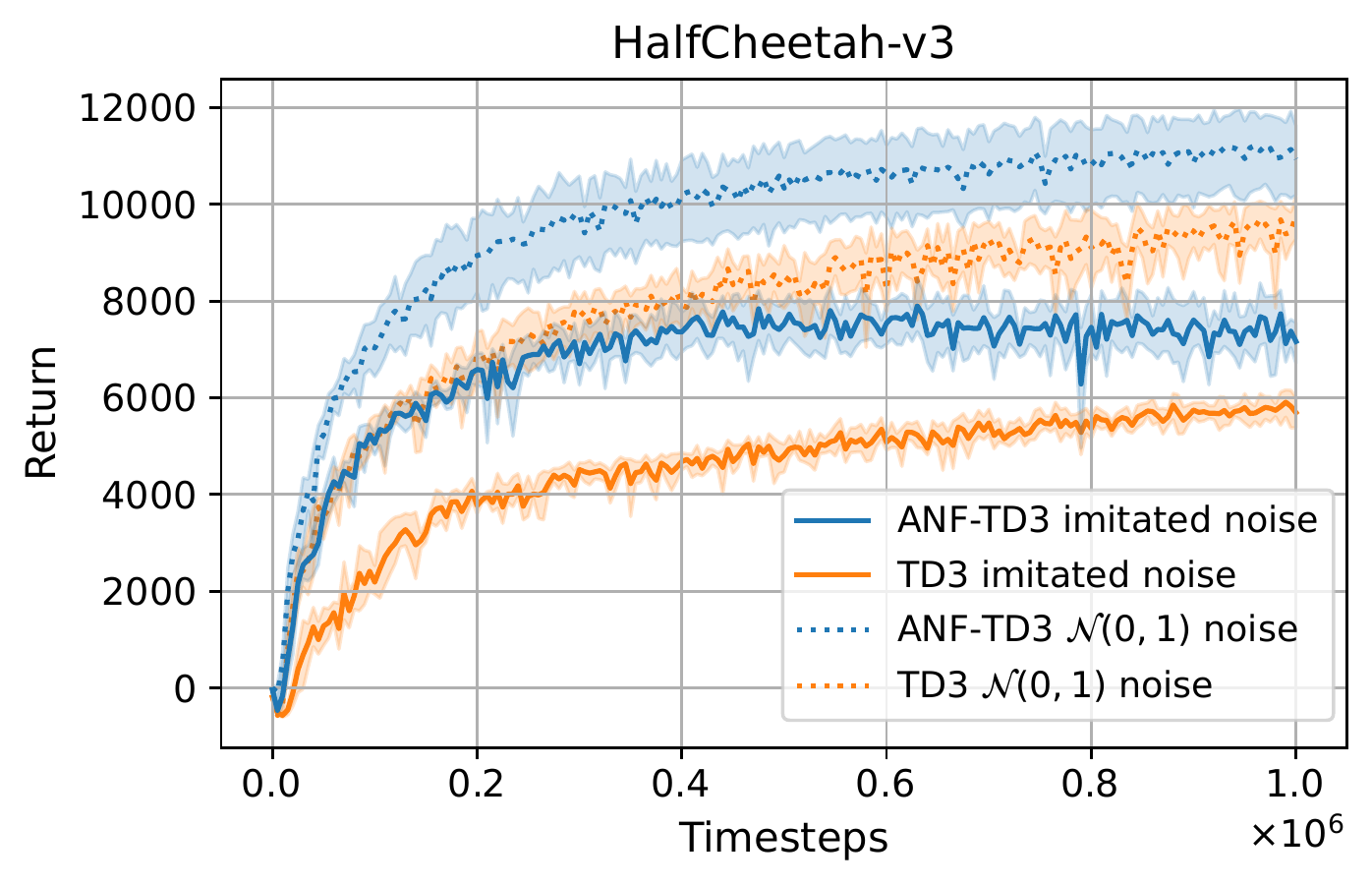}
    \vspace{-0.4cm}
    \caption{Learning curves for ANF-TD3 and its dense baseline on the challenging ENE, where the noise features imitate the task-relevant features. This increases the difficulty, but ANF still achieves the highest return. The ENE has 90\% of these \mbox{realistic} noise features in this experiment. See Figure~\ref{fig:imitated_HalfCheetah-curves-SAC} in Appendix~\ref{sec:app-imitate} for the graph of ANF-SAC.}
    \vspace{-0.2cm}
    \label{fig:imitated_HalfCheetah-curves-TD3}
\end{figure}

\textbf{\textit{Results.}}
In Figure~\ref{fig:imitated_HalfCheetah-curves-TD3}, we see that the imitated noise indeed raises the difficulty of the ENE. The performance of both ANF-TD3 and TD3 decreases considerably compared to the standard $\mathcal{N}(0,1)$ noise. However, ANF is still able to outperform its dense baseline by a large margin, even in this challenging ENE.

\section{Does ANF need to be dynamic?}
\label{sec:static}

In this section, we perform an ablation study to show that the dynamic network topology updates of ANF are a necessary component of the algorithm. Removing these dynamic updates during training would give a static sparse training algorithm. This \mbox{algorithm} starts with a randomly sparsified input layer just like ANF, but it does not drop or regrow any connections during training.

\textbf{\textit{Experimental setup.}}
We compare ANF to its fixed-connectivity counterpart, which we call \textit{Static-ANF}. 
In addition, we compare the standard dense algorithms of SAC and TD3.
We run on Humanoid-v3 with 90\% noise features.

\begin{figure}
    \centering
    \includegraphics[width=\linewidth]{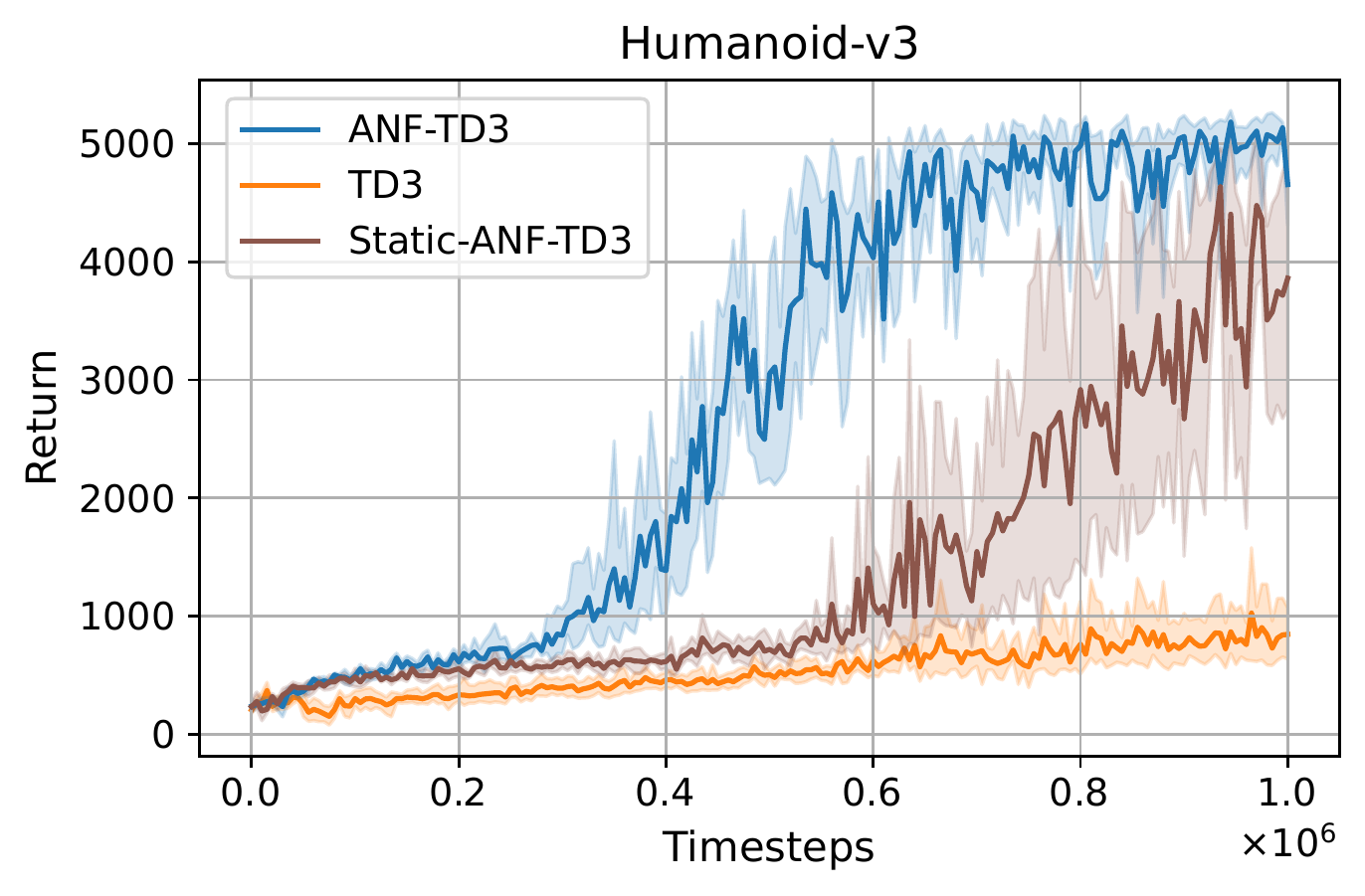}
    \vspace{-0.8cm}
    \caption{Comparison of ANF to its static sparse counterpart and standard (fully dense) TD3.
    ANF-TD3 dynamically updates the sparse network topology, while the topology of the other two methods remains static. The learning curves show that the dynamic updates are an essential part of ANF.
    This experiment is run with 90\% noise features. A similar graph for SAC is shown in Figure~\ref{fig:static_curves_fake90_Humanoid_SAC} of Appendix~\ref{sec:app-static}.}
    \vspace{-0.4cm}
    \label{fig:static_curves_fake90_Humanoid_TD3}
\end{figure}

\textbf{\textit{Results.}}
The graphs in Figure~\ref{fig:static_curves_fake90_Humanoid_TD3} show that ANF indeed needs to be dynamic, as it significantly outperforms its static version. Intuitively, this is consistent with the concept shown in Figure~\ref{fig:connectivity_over_time_HalfCheetah_TD3_critic1}, where ANF changes its connectivity to emphasize its focus on the task-relevant features. This emphasis is lost if one removes ANF's ability to dynamically adjust the network structure.

Nevertheless, it is remarkable to see that Static-ANF is able to surpass standard dense TD3 in this setting. It seems that just having fewer connections to the 90\% noisy input features already helps to lower the overall distraction.

\section{How Sparse can we go?}
\label{sec:sparse}



We already showed that sparsifying the input layer can significantly improve performance in extremely noisy environments.
In this section, we investigate whether we can further sparsify the agent's networks by also pruning connections in other layers.
This would reduce the network size (total number of parameters) even further, while hopefully maintaining performance.


\textbf{\textit{Experimental setup.}}
Instead of only having an 80\% sparse \textit{input} layer, the networks now also have a sparse \textit{hidden} layer for which the connectivity is frequently adjusted with DST. We keep the output layer dense, just as in \cite{ghada}.
The sparsity distribution is uniform, meaning that both the input layer and the hidden layer have the same sparsity level. We compute the required layer sparsity levels such that the global sparsity level (over the full network) is at $s$, where $s$ ranges over $\{.80, .90, .95, .98\}$. 
We compare with our standard ANF algorithm, which has a global sparsity of 74.6\%  for \mbox{$n_f\!=\!0.9$} (actor network on Humanoid).
We run on Humanoid-v3 and HalfCheetah-v3, as the achievable sparsity level before performance degradation differs significantly between these two environments.



\begin{figure}
    \centering
    \includegraphics[width=\linewidth]{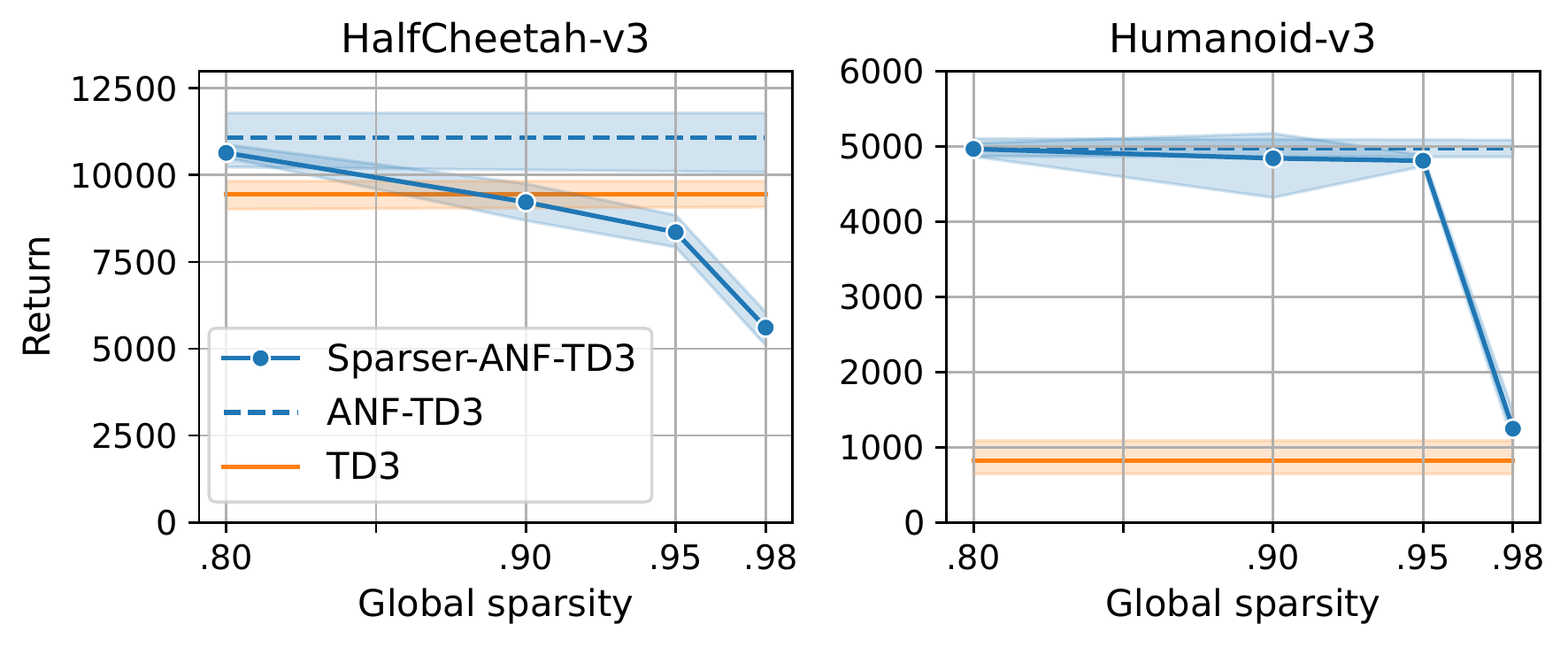}
    \vspace{-0.8cm}
    \caption{Performance of ANF-TD3 and its sparser versions, on 90\% noise features. Sparser-ANF also prunes weights in the hidden layer, instead of only the input layer. Further sparsifying the network does not improve performance, but can drastically reduce the size of the network. The graphs for ANF-SAC are shown in Figure~\ref{fig:sparse_2envs_SAC} of Appendix~\ref{sec:app-sparser}.}
    \vspace{-0.2cm}
    \label{fig:sparse_2envs_TD3}
\end{figure}

\textbf{\textit{Results.}}
We see in Figure~\ref{fig:sparse_2envs_TD3} that on extremely noisy environments, nearly the same performance can be reached with further sparsified networks, meaning that a large proportion of the parameters can be pruned.
In Figure~\ref{fig:sparse_2envs_TD3} (left), we see that ANF-TD3 with a global sparsity of 80\% still surpasses standard dense TD3 in final return on HalfCheetah-v3. As the global sparsity increases, the performance gradually decreases.
This is quite different for Humanoid-v3, in Figure~\ref{fig:sparse_2envs_TD3} (right). Notice that ANF-TD3 can go up to a global sparsity level of 95\%, with barely any performance degradation. This means it can use $20\times$ fewer parameters than standard TD3, reducing the network size considerably, as shown in Table~\ref{tab:overview}. 

\begin{table}
\caption{Comparison of different algorithms, along with their parameter counts for the actor network. The critic networks have comparable numbers of parameters.}
\label{tab:overview}
\vspace{-0.2cm}
\setlength{\tabcolsep}{4pt}
\resizebox{\columnwidth}{!}{%
\begin{tabular}{llrr}\toprule
Algorithm             & Environment     & Return ($\uparrow$) & \# Params. ($\downarrow$) \\ 
\midrule
ANF-TD3               & Humanoid     & \textbf{4968.3}   & 262,400     \\ 
TD3                   & Humanoid     & 817.3             & 1,032,448   \\ 
Sparser(80\%)-ANF-TD3 & Humanoid     & 4963.1            & 206,489     \\ 
Sparser(95\%)-ANF-TD3 & Humanoid     & 4806.4            & \textbf{51,622}      \\ 
ANF-TD3               & HalfCheetah  & \textbf{11086.4}  & 75,776      \\ 
TD3                   & HalfCheetah  & 9452.6            & 110,592     \\ 
Sparser(80\%)-ANF-TD3 & HalfCheetah  & 10640.3           & 22,118      \\ 
Sparser(95\%)-ANF-TD3 & HalfCheetah  & 8357.9            & \textbf{5,529}       \\ 
\bottomrule         
\end{tabular}%
}
\vspace{-0.35cm}
\end{table}

\section{Conclusion \& Limitations}
\label{sec:conclusion}

In this work, we formulated the problem setting of extremely noisy environments and showed that our Automatic Noise Filtering algorithm succeeds at this challenge where standard deep RL methods struggle. By using dynamic sparse training, the ANF algorithm adjusts its network topology to focus on task-relevant features.

Our experiments provide an initial empirical verification of our Adaptability Hypothesis, which roughly states that dropping and growing sparse connections is easier than adjusting dense weights. Further research is necessary to grant more conclusive evidence, as our work is limited to continuous control tasks that have feature vectors as states. We exclusively studied SAC and TD3; integrating ANF in other deep RL methods is open for future work.

The input layer sparsity is an important hyperparameter for ANF. Making this an adaptive parameter could be beneficial.
Further, we believe expanding ANF's compatibility towards other neural network types is a promising future research direction. Combining ANF with convolutional NNs or transformers would open the possibility of operating on noisy image or video data.





\begin{acks}
This publication is part of the project AMADeuS (with project number 18489)
of the Open Technology Programme, which is partly financed by the Dutch Research Council (NWO).
This research used the Dutch national e-infrastructure with the support of the SURF Cooperative, using grant no. EINF-3098.
Part of this work has taken place in the Intelligent Robot Learning (IRL) Lab at the University of Alberta, which is supported in part by research grants from the Alberta Machine Intelligence Institute (Amii); a Canada CIFAR AI Chair, Amii; Compute Canada; Huawei; Mitacs; and NSERC.
We thank Joan Falc\'o Roget, Mickey Beurskens, Anne van den Berg, and Rik Grooten for the fruitful discussions. Finally, we thank the anonymous reviewers and Antonie Bodley for their thorough proofreading and useful comments.
\end{acks}



\bibliographystyle{ACM-Reference-Format} 
\balance
\bibliography{references}


\newpage
\onecolumn
\section*{Appendix}

\appendix

\vspace{1em}
\section{ANF algorithm}
\label{sec:app-pseudocode}

In this section we provide pseudocode of the Automatic Noise Filtering (ANF) algorithm. In Algorithm~\ref{alg:ANF_SAC} we show the implementation of ANF-SAC, but keep in mind that ANF can be applied to any MLP-based deep RL method.
The novel parts of our proposed method ANF are colored {\color{xkcdViolet}violet}. The dynamic sparse training components, already introduced by \cite{ghada}, are colored {\color{xkcdMediumBlue}medium-blue}.\footnote{Colors from \url{https://xkcd.com/color/rgb/}.} The rest (in black) is the standard SAC algorithm \cite{sac}.

Our code is open-source and can be found at: \url{https://github.com/bramgrooten/automatic-noise-filtering}.


\vspace{1em}

{\centering
\begin{minipage}{.8\linewidth}
\begin{algorithm}[H]
\caption{ANF-SAC}
\label{alg:ANF_SAC}
\begin{algorithmic}[1]
\REQUIRE {\color{xkcdViolet}input layer sparsity $s_i$}, {\color{xkcdMediumBlue}topology-change period $\Delta T$, drop fraction $d_f$}, initial collect steps $b_{init}$, train every $k$ env steps, minibatch size $n$, learning rate $\lambda$, target smoothing coefficient $\tau$, max env steps $T$
\STATE Initialize the actor network $\pi$ and two critic networks $Q_1$, $Q_2$, with weights $\phi, \theta_1, \theta_2$.
\STATE {\color{xkcdViolet}Randomly prune the input layer of all 3 networks to sparsity level $s_i$.}
\STATE Duplicate the two critics to create two target networks, with weights $\bar{\theta}_1 = \theta_1$, $\bar{\theta}_2 = \theta_2$.
\STATE Initialize the replay buffer $\mathcal{B}$ with $b_{init}$ random actions.
\FOR{$t=1$ {\bfseries to} $T$}
    \STATE Sample action $a$ from the policy (actor network) based on current state $s$: $a \sim \pi_{\phi}(\cdot | s)$
    \STATE Take a step in the environment and observe reward $r$ and new state $s'$.
    \STATE Store transition tuple $(s, a, r, s')$ in $\mathcal{B}$.
    \IF{$t$ mod $k == 0$} 
        \STATE Sample minibatch of $n$ transitions from $\mathcal{B}$.
        \STATE Update the weights according to SAC's objective functions $J_Q$, $J_{\pi}$:
        \STATE \ \ \ \ $\theta_i \leftarrow \theta_i - \lambda \hat{\nabla}_{\theta_i} J_Q(\theta_i)$ \ for $i \in \{1,2\}$
        \STATE \ \ \ \ $\phi \leftarrow \phi - \lambda \hat{\nabla}_{\phi} J_{\pi}(\phi)$
        \STATE Update the target networks: 
        \STATE \ \ \ \ $\bar{\theta_i} \leftarrow \tau \theta_i + (1 - \tau) \bar{\theta_i}$ \ for $i \in \{1,2\}$
    \ENDIF
    {\color{xkcdMediumBlue}
    \IF{$t$ mod $\Delta T == 0$}
        \STATE Update the topology of the networks:
        \STATE \ \ \ \ Prune fraction $d_f$ of the smallest magnitude weights.
        \STATE \ \ \ \ Grow fraction $d_f$ new weights randomly, initialize at value 0.
        {\color{xkcdViolet}
        \STATE \ \ \ \ Mask Adam's running avg. of the 1st \& 2nd raw moment of the gradient for pruned weights: 
        \STATE \ \ \ \ \ \ \ \ $m \leftarrow 0$, $v \leftarrow 0$.    
        }
    \ENDIF
    }
\ENDFOR
\end{algorithmic}
\end{algorithm}
\end{minipage}
\par
}

\vspace{2em}

\section{Experimental Details}
\label{sec:app-hyper}

In this section we present an overview of the hyperparameters used in our experiments.

\textbf{Algorithms:} Throughout the experiments in the paper we compared our proposed dynamic sparse algorithms (ANF-SAC and ANF-TD3) with the fully-dense counterparts (SAC and TD3), and the static sparse variants (Static-ANF-SAC and Static-ANF-TD3). The complete list of hyperparameters can be found in Table~\ref{tab:hyper}.
Aiming for a fair comparison, we tried to maximize the number of shared hyperparameters. 
Table~\ref{tab:hyper} also includes the parameters used for the experiments described in Section~\ref{sec:sparse}, where an increasing sparsity level is incorporated in ANF beyond the first layer, leading to the Sparser-ANF-SAC and Sparser-ANF-TD3 versions. 

\begin{figure}[H]
    \centering
    \includegraphics[width=0.75\textwidth]{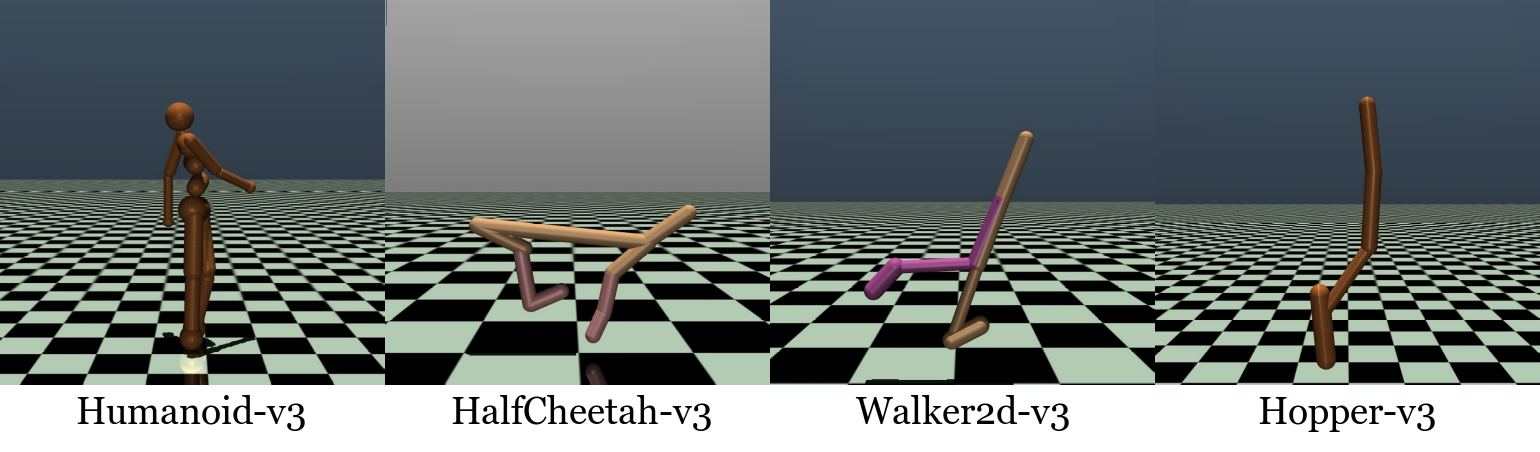}
    \vspace{-0.3cm}
    \caption{The MuJoCo Gym \cite{mujoco, gym} environments used as a base for our ENEs.}
    \label{fig:mujoco_gym_4envs_captions}
\end{figure}

\textbf{Environments:} We used as the foundation for our extremely noisy environments (ENEs) four continuous control tasks (Humanoid-v3,\footnote{Please note that according to the environment's documentation (\url{https://www.gymlibrary.dev/environments/mujoco/humanoid/\#observation-space}) versions 1,2,3 of Humanoid contain an issue with the contact forces (the corresponding features always give 0). Humanoid-v4 solves this, but came out too late for our research.} HalfCheetah-v3, Walker-v3, and Hopper-v3) as shown in Figure~\ref{fig:mujoco_gym_4envs_captions}. See Table~\ref{tab:ene} for the parameters corresponding to the ENEs built on top of these four worlds.

\begin{table}[H]
\caption{Hyperparameters. Table format from \cite{sac}.}
\label{tab:hyper}
\vspace{-0.2cm}
\begin{tabular}{ll}\toprule
Parameter                                             & Value      \\ \midrule
\textit{Shared by all algorithms}                     &             \\ 
\quad optimizer                                       & Adam \cite{adam}    \\ 
\quad learning rate ($\lambda$)                       & $1\cdot 10^{-3}$  \\
\quad weight decay                                    & $2\cdot 10^{-4}$  \\
\quad discount ($\gamma$)                             & $0.99$               \\ 
\quad nonlinearity                                    & ReLU            \\ 
\quad replay buffer size                              & $10^6$    \\
\quad initial collect steps ($b_{init}$)              & $25,000$  \\
\quad network type                                    & MLP   \\
\quad number of hidden layers                         & $2$  \\
\quad number of neurons per hidden layer              & $256$  \\
\quad minibatch size ($n$)                            & $100$   \\
\quad target smoothing coefficient ($\tau$)           & $0.005$  \\
\quad train every $k_c$ env steps (critic), $k_c=$    & $1$   \\ 
\quad gradient steps per training step $=$            & $1$     \\
\midrule
\textit{SAC and ANF-SAC}                              &         \\ 
\quad SAC type (Gaussian / Deterministic)             & Gaussian \\
\quad temperature ($\alpha$ in \cite{sac})            & $0.2$ \\
\quad automatic temperature tuning                    & False \\
\quad train every $k_a$ env steps (actor), $k_a=$     & $1$   \\
\quad target update interval ($k_{tar}$)              & $1$    \\  
\midrule
\textit{TD3 and ANF-TD3}                              &         \\ 
\quad train every $k_a$ env steps (actor), $k_a=$     & $2$   \\
\quad target update interval ($k_{tar}$)              & $2$    \\ 
\quad std. dev. of exploration noise ($\sigma$ in \cite{td3})  & $0.1$  \\
\quad std. dev. of sampling noise ($\widetilde{\sigma}$ in \cite{td3})  & $0.2$  \\
\quad sampling noise clip ($c$ in \cite{td3})         & $0.5$  \\
\midrule
\textit{ANF}                                          &  \\
\quad sparsity of the input layer ($s_i$)             & $0.8$ \\
\quad drop fraction ($d_f$)                           & $0.05$ \\
\quad new weights init value                          & $0$  \\
\quad sparse layers                                   & input layer  \\
\quad topology-change period ($\Delta T$), env steps  & $1000$ \\
\midrule
\textit{Static-ANF}                                   &  \\
\quad topology-change period ($\Delta T$), env steps  & $\infty$ (no change) \\
\midrule
\textit{Sparser-ANF}                                  &  \\
\quad global sparsity                                 & varying (Section~\ref{sec:sparse})  \\
\quad sparsity distribution (over sparse layers)      & uniform  \\
\quad sparse layers                                   & input \& hidden  \\
\bottomrule         
\end{tabular}
\end{table}

\begin{table}[H]
\caption{Extremely noisy environment (ENE) parameters.}
\label{tab:ene}
\vspace{-0.2cm}
\begin{tabular}{lll}\toprule
Parameter                        & Value      & Experiments in   \\ \midrule
noise fraction ($n_f$)           & varying    & Section~\ref{sec:anf}    \\ 
noise distribution               & $\mathcal{N}(0,\sigma^2)$ & Sections~\ref{sec:anf}-\ref{sec:louder}, \ref{sec:static}-\ref{sec:sparse}  \\    
                                 & Imitate  & Section~\ref{sec:imitate} \\
noise amplitude ($\sigma$)       & $1$ & Sections~\ref{sec:anf}-\ref{sec:transfer}, \ref{sec:static}-\ref{sec:sparse} \\
                                 & varying  & Section~\ref{sec:louder} \\
permutation period ($T_p$)       & 1M   & Section~\ref{sec:transfer} \\
\bottomrule         
\end{tabular}
\end{table}

\vspace{2cm}

\section{Sparse Hardware and Software support}
\label{sec:app-hardware}

As discussed in \cite{ghada}, research on sparsity is moving in three simultaneous directions as a collaborative community effort. Firstly, hardware that can benefit from sparse neural networks. In order to support a sparsity level of 50\% as a first step, NVIDIA produced the A100 \cite{zhou}. Secondly, software libraries that support neural network implementations which are truly sparse. Supervised learning has begun to receive attention in this direction \cite{shiwei_million, curci2021truly}. Thirdly, algorithmic approaches, which are the subject of our research, are intended to offer sparse network methods whose performance is at least at the level of dense models \cite{hoefler}.
We will be able to produce faster, memory-efficient, and energy-efficient deep neural networks with parallel efforts in the three dimensions. Further discussion of this can be found in \cite{hardware_lottery, mocanu_aamas}.


\newpage

\section{Additional Results}

In this appendix we share results that did not fit into the main body of the paper. Many figures present outcomes of additional algorithms or environments, while some graphs such as Figure~\ref{fig:connections_actor_HalfCheetah} and \ref{fig:permuted_actor_HalfCheetah} show extra analysis of the network connectivity.

\subsection{Automatic Noise Filtering graphs}
\label{sec:app-anf}

\begin{figure}[H]
    \centering
    \captionsetup{width=0.98\textwidth}
    \includegraphics[width=\textwidth]{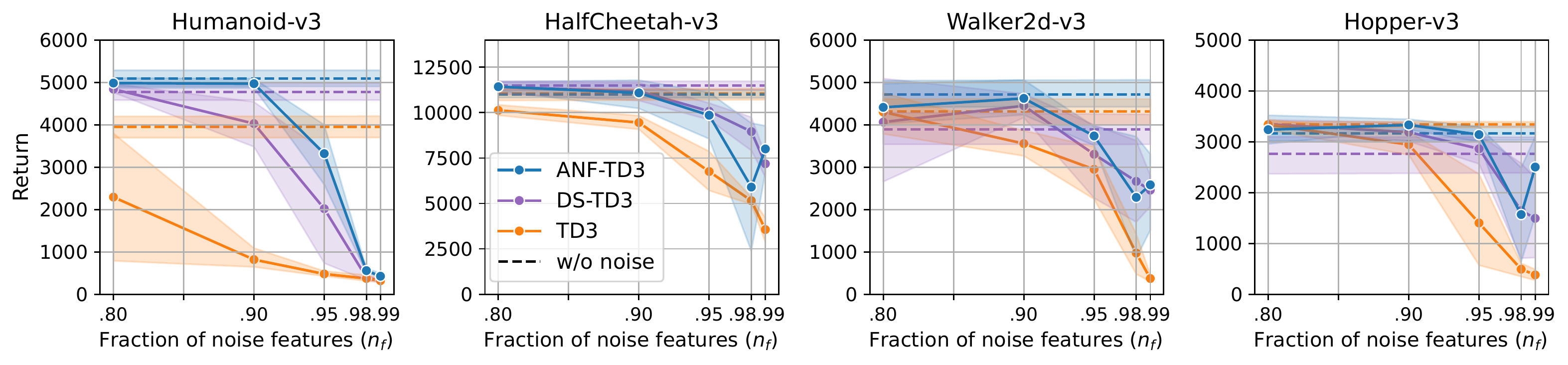}
    \caption{Performance of ANF compared to its baselines for different fractions of noise features $n_f$. When the environments contain a lot of noise (high $n_f$) the standard dense networks of TD3 seem to struggle. Similar graphs for ANF-SAC are shown in Figure~\ref{fig:overview80-99_horiline_SAC} of Section~\ref{sec:anf}.}
    \label{fig:overview80-99_horiline_TD3}
\end{figure}

Additional results on the main ANF experiments are shown in Figure~\ref{fig:overview80-99_horiline_TD3}.

\begin{figure}[H]
    \centering
    \captionsetup{width=0.6\textwidth}
    \includegraphics[width=0.5\textwidth]{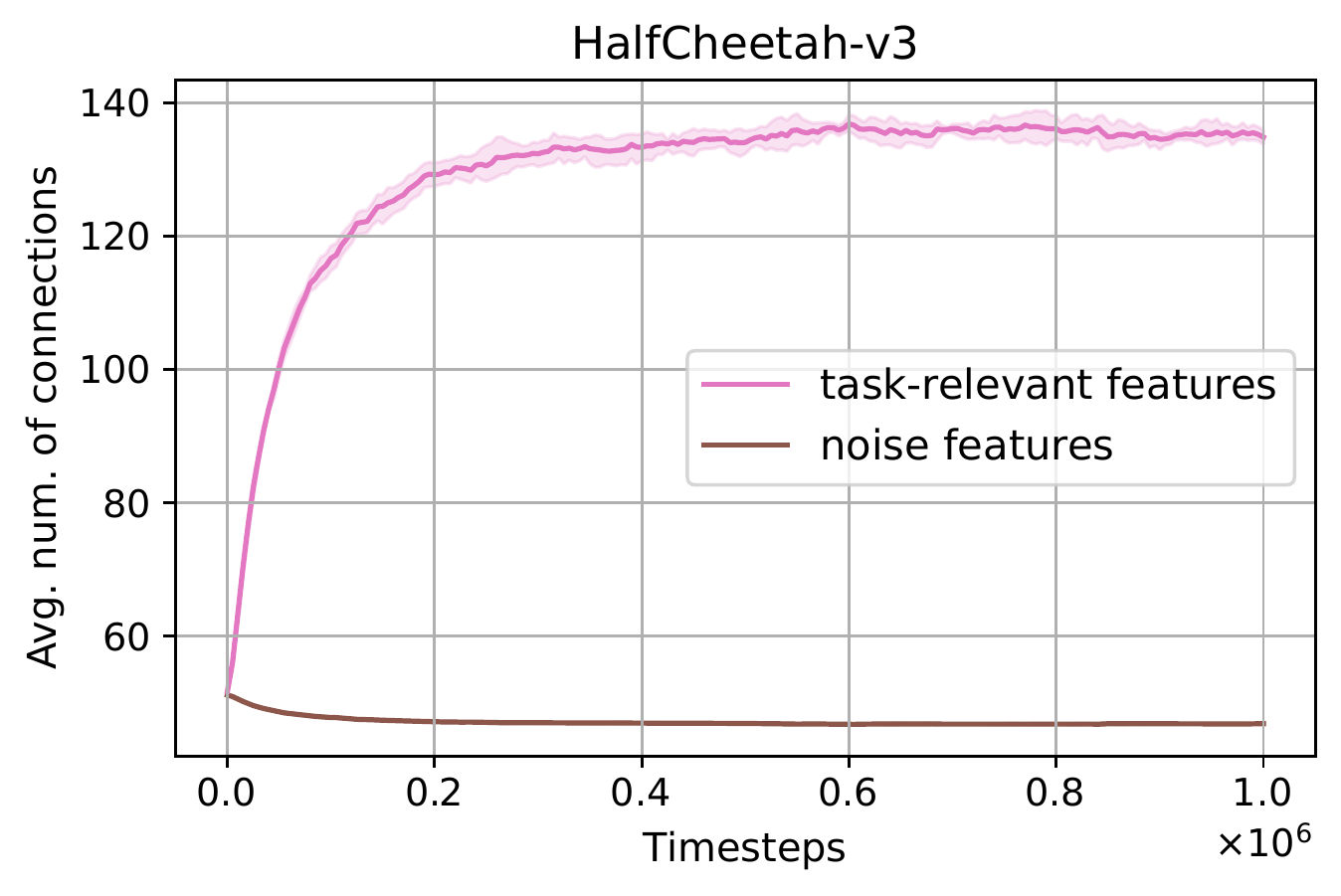}
    \caption{Average number of connections in the input layer of one of ANF-SAC's critic networks, on HalfCheetah-v3 with 90\% noise features. At the start of training every input neuron has around $256 \cdot 0.2 \approx 51$ connections, because the input layer sparsity is 80\% and connections are allocated uniformly at random. During training, ANF gradually prunes connections from the noise features and grows connections to the relevant features. A similar graph for ANF-TD3 is shown in Figure~\ref{fig:connectivity_over_time_HalfCheetah_TD3_critic1} of Section~\ref{sec:anf}.}
    \label{fig:connectivity_over_time_HalfCheetah_SAC_critic1}
\end{figure}

\begin{figure}[H]
    \centering
    \captionsetup{width=0.6\textwidth}
    \includegraphics[width=0.6\textwidth]{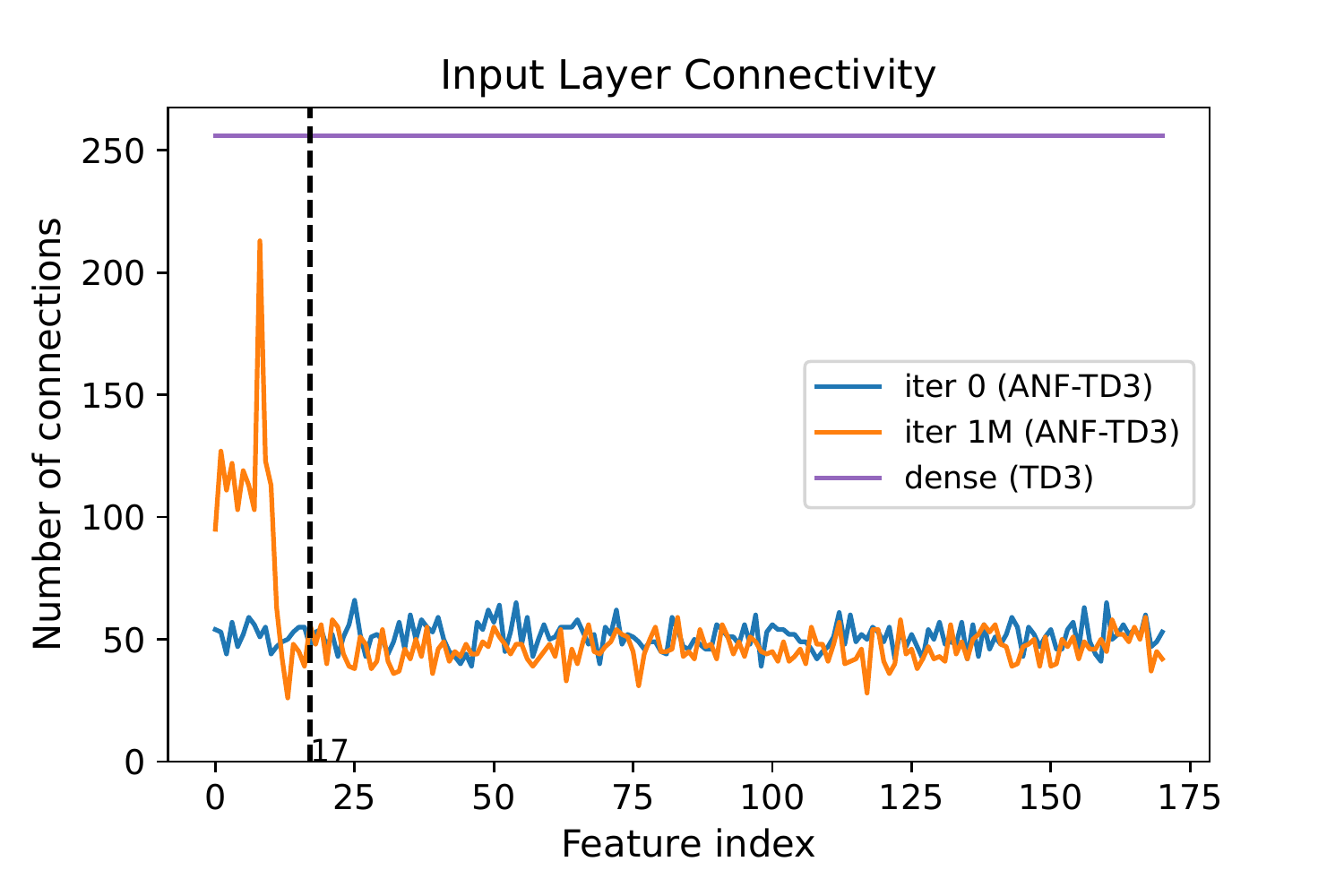}
    \caption{Showing the connectivity of the actor network's input layer for ANF-TD3 on HalfCheetah-v3 with 90\% noise features. At the start of training (iteration 0) every input neuron has around $256 \cdot 0.2 \approx 51$ connections, because the input layer sparsity is 80\% and connections are allocated uniformly at random. During training, ANF-TD3 gradually removes connections from the noise features and adds connections to the relevant features (the 17 leftmost input neurons in this graph). Note that a dense network always has 256 connections to every input neuron, and thus gets distracted more easily by all the noise features.}
    \label{fig:connections_actor_HalfCheetah}
\end{figure}

A noteworthy observation for Figure~\ref{fig:connections_actor_HalfCheetah} is the fact that apparently not all 17 input features are deemed to be useful, as some received less than the original 51 connections. This phenomenon has been shown in standard RL environments recently by \cite{mintaskrepr}, and it is interesting to see that it still holds up in extremely noisy environments. Original state features that fall outside the \textit{minimal task representation} are considered just as irrelevant as the noise features, according to Figure~\ref{fig:connections_actor_HalfCheetah}. We see that ANF filters not only through synthetic features, but also de-emphasizes redundant features.

The full version of Table~\ref{tab:dims} with all noise fractions used is shown below in Table~\ref{tab:dims-full}.

\begin{table}[H]
\caption{State and action space dimensions.}
\label{tab:dims-full}
\vspace{-0.2cm}
\setlength{\tabcolsep}{4pt}
\begin{tabular}{llllllll}\toprule
Environment    & State dim. & Action dim. & State dim.  & State dim. & State dim. & State dim. & State dim. \\
               & \small{\textit{Original}} & \small{\textit{Original}} & \small{\textit{ENE ($n_f\!=\!.8$)}} & \small{\textit{ENE ($n_f\!=\!.9$)}} & \small{\textit{ENE ($n_f\!=\!.95$)}}  & \small{\textit{ENE ($n_f\!=\!.98$)}}   & \small{\textit{ENE ($n_f\!=\!.99$)}}  \\  \midrule
Humanoid-v3    & 376        & 17          & 1880     & 3760  & 7520  & 18,800  & 37,600        \\ 
HalfCheetah-v3 & 17         & 6           & 85       & 170   & 340   & 850 & 1700        \\ 
Walker2d-v3    & 17         & 6           & 85       & 170   & 340   & 850 & 1700        \\ 
Hopper-v3      & 11         & 3           & 55       & 110   & 220   & 550 & 1100        \\ \bottomrule         
\end{tabular}%
\end{table}

\newpage
\subsection{Transfer Learning graphs}
\label{sec:app-transfer}

\begin{figure}[H]
    \centering
    \captionsetup{width=0.8\textwidth}
    \includegraphics[width=0.75\textwidth]{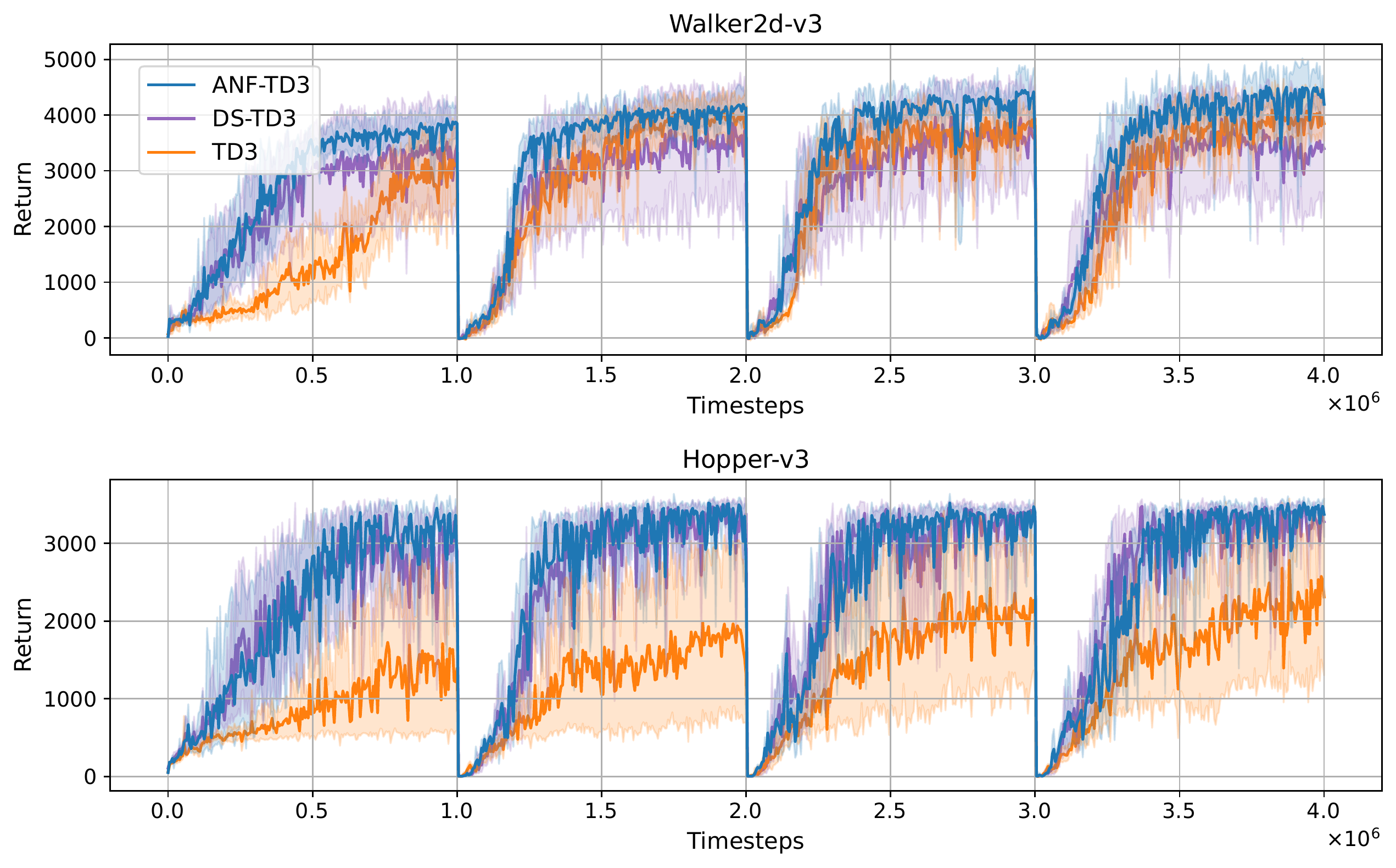}
    \vspace{-0.35cm}
    \caption{Performance of ANF-TD3 and its baselines on permuted extremely noisy environments (PENE) with 95\% noise features. After every 1M timesteps the features are shuffled with a random permutation. ANF is able to cope with this challenge better than the fully dense networks of TD3.}
    \label{fig:transfer_WalkerHopper_TD3}
\end{figure}

\begin{figure}[H]
    \centering
    \captionsetup{width=0.8\textwidth}
    \includegraphics[width=0.75\textwidth]{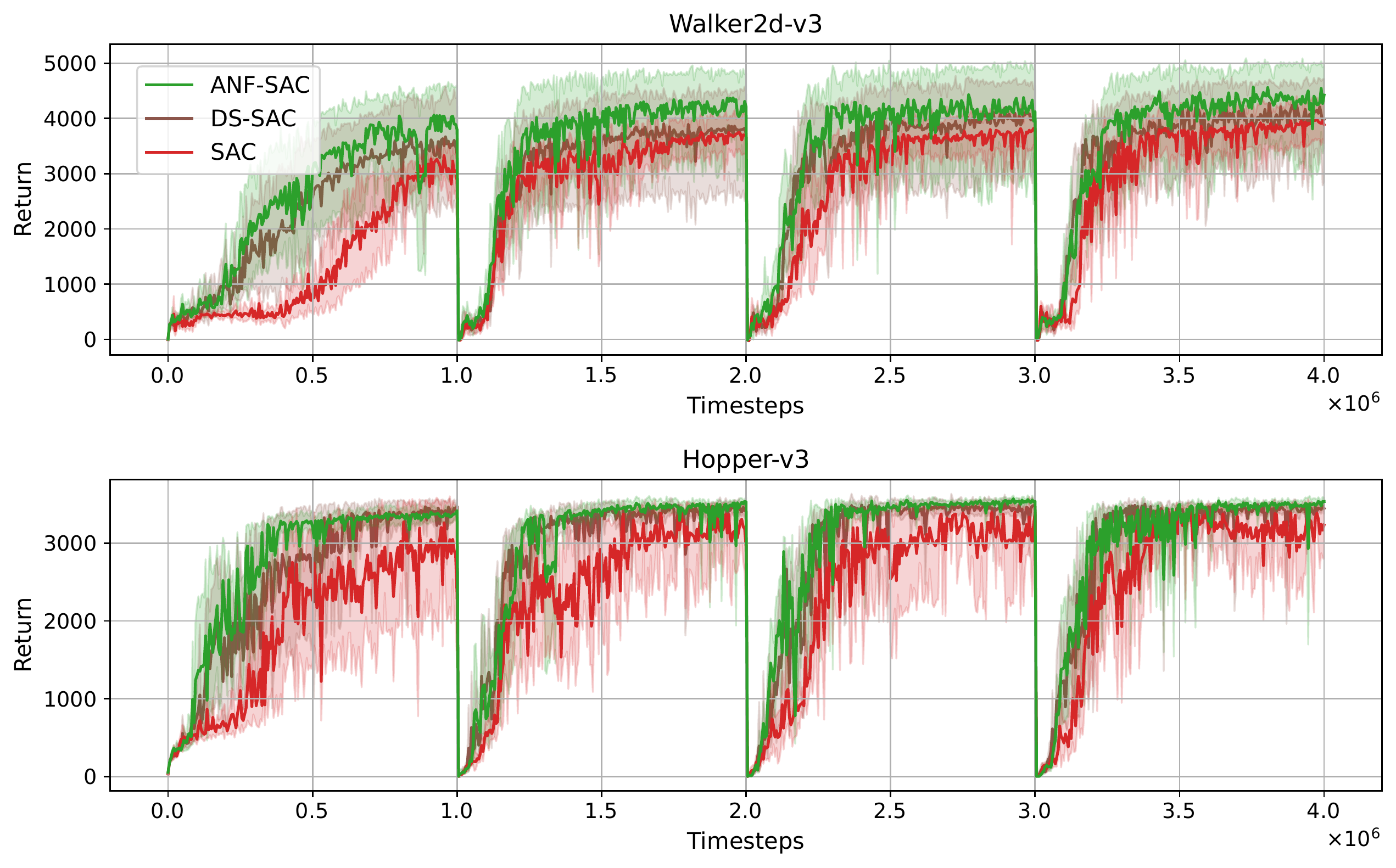}
    \vspace{-0.35cm}
    \caption{Performance of ANF-SAC and its baselines on permuted extremely noisy environments (PENE) with 95\% noise features. After every 1M timesteps the features are shuffled with a random permutation. ANF is able to cope with this challenge better than the fully dense networks of SAC.}
    \label{fig:transfer_WalkerHopper_SAC}
\end{figure}

\begin{figure}[H]
    \centering
    \captionsetup{width=0.83\textwidth}
    \includegraphics[width=0.75\textwidth]{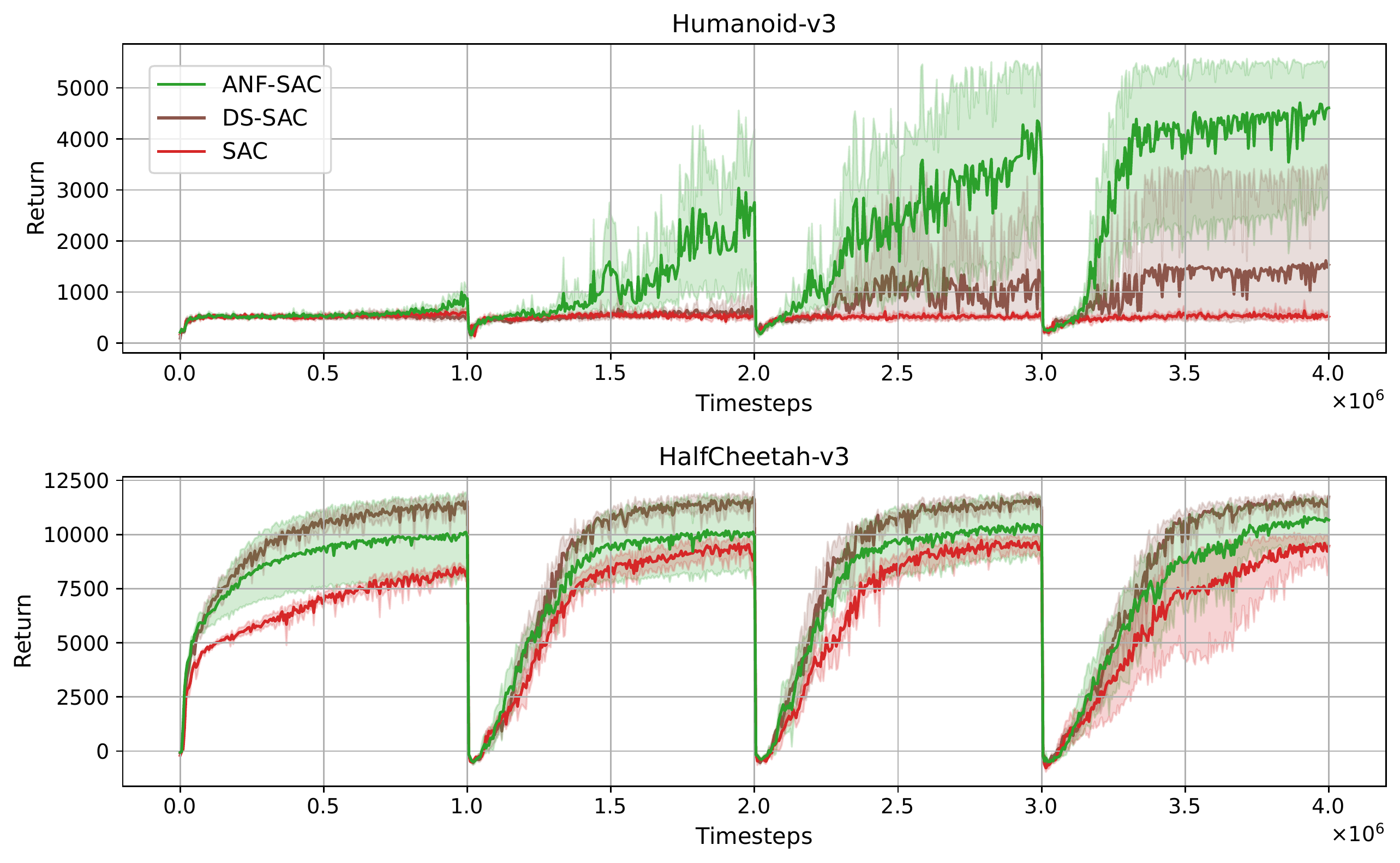}
    \vspace{-0.35cm}
    \caption{Performance of ANF-SAC and its baselines on permuted extremely noisy environments (PENE) with 95\% noise features. After every 1M timesteps the features are shuffled with a random permutation. ANF is able to cope with this challenge, while the fully dense networks of SAC are struggling.}
    \label{fig:transfer_HumanoidHalfCheetah_SAC}
\end{figure}

\begin{figure}[H]
    \centering
    \captionsetup{width=0.6\textwidth}
    \includegraphics[width=0.6\textwidth]{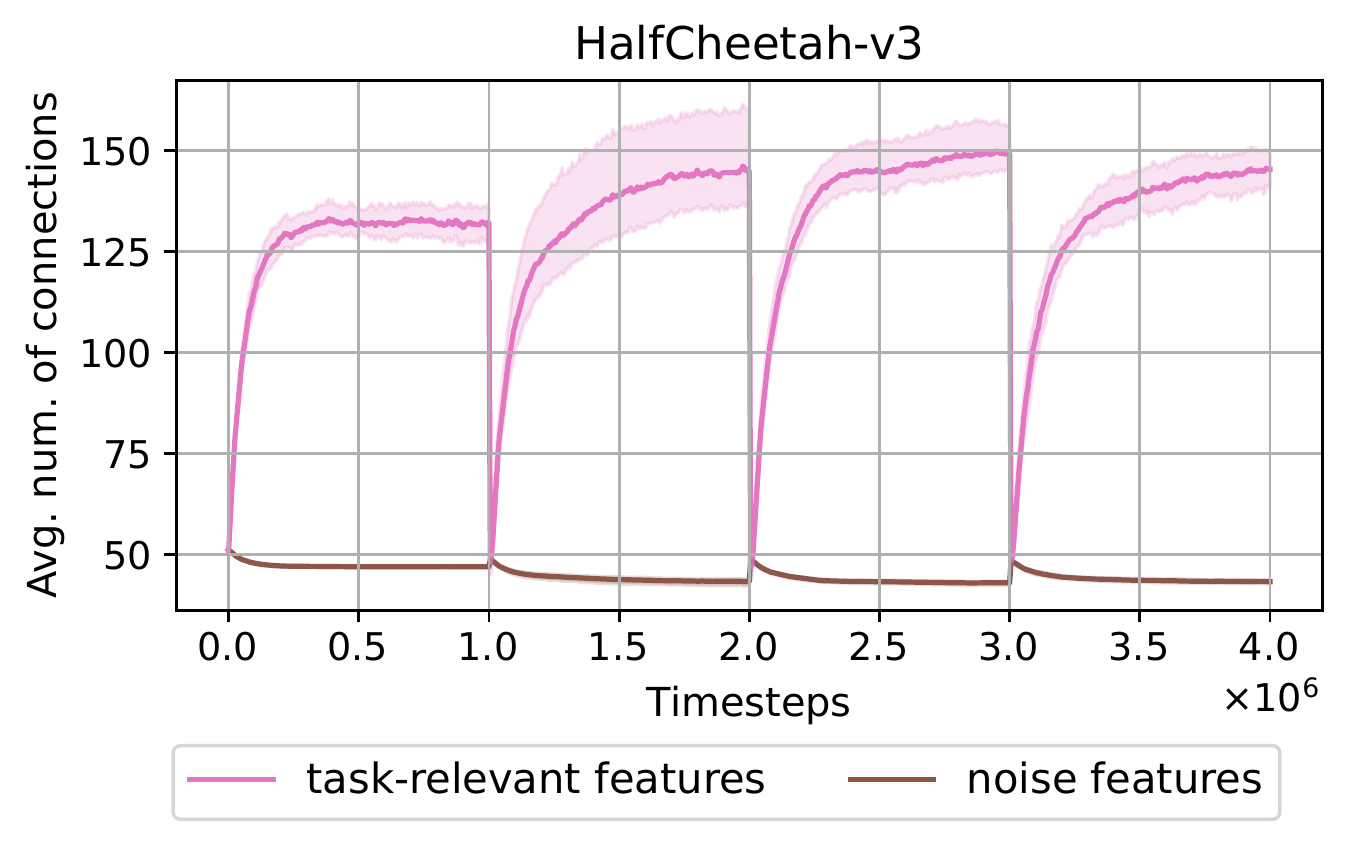}
    \caption{Average number of connections in the input layer of one of ANF-TD3's critic networks, on HalfCheetah-v3 with 95\% noise features. 
    After every 1M timesteps the PENE rearranges the order of the features by a random permutation. The ANF agent adjusts its network structure quickly, growing connections with the task-relevant features which are now coming in at different input neurons. 
    A similar graph for ANF-SAC is shown in Figure~\ref{fig:connectivity_over_4Mtime_HalfCheetah_SAC_critic1} of Section~\ref{sec:transfer}.
    }
    \label{fig:connectivity_over_4Mtime_HalfCheetah_TD3_critic1}
\end{figure}

Extra graphs on transfer learning shown in Figures~ \ref{fig:transfer_WalkerHopper_TD3}, \ref{fig:transfer_WalkerHopper_SAC}, \ref{fig:transfer_HumanoidHalfCheetah_SAC}. They correspond to Figure~\ref{fig:transfer_HumanoidHalfCheetah_TD3} in Section~\ref{sec:transfer} of the paper. 
The connectivity graph of ANF-TD3 is given in Figure~\ref{fig:connectivity_over_4Mtime_HalfCheetah_TD3_critic1}.

\begin{figure}[H]
    \centering
    \captionsetup{width=0.6\textwidth}
    \includegraphics[width=0.6\textwidth]{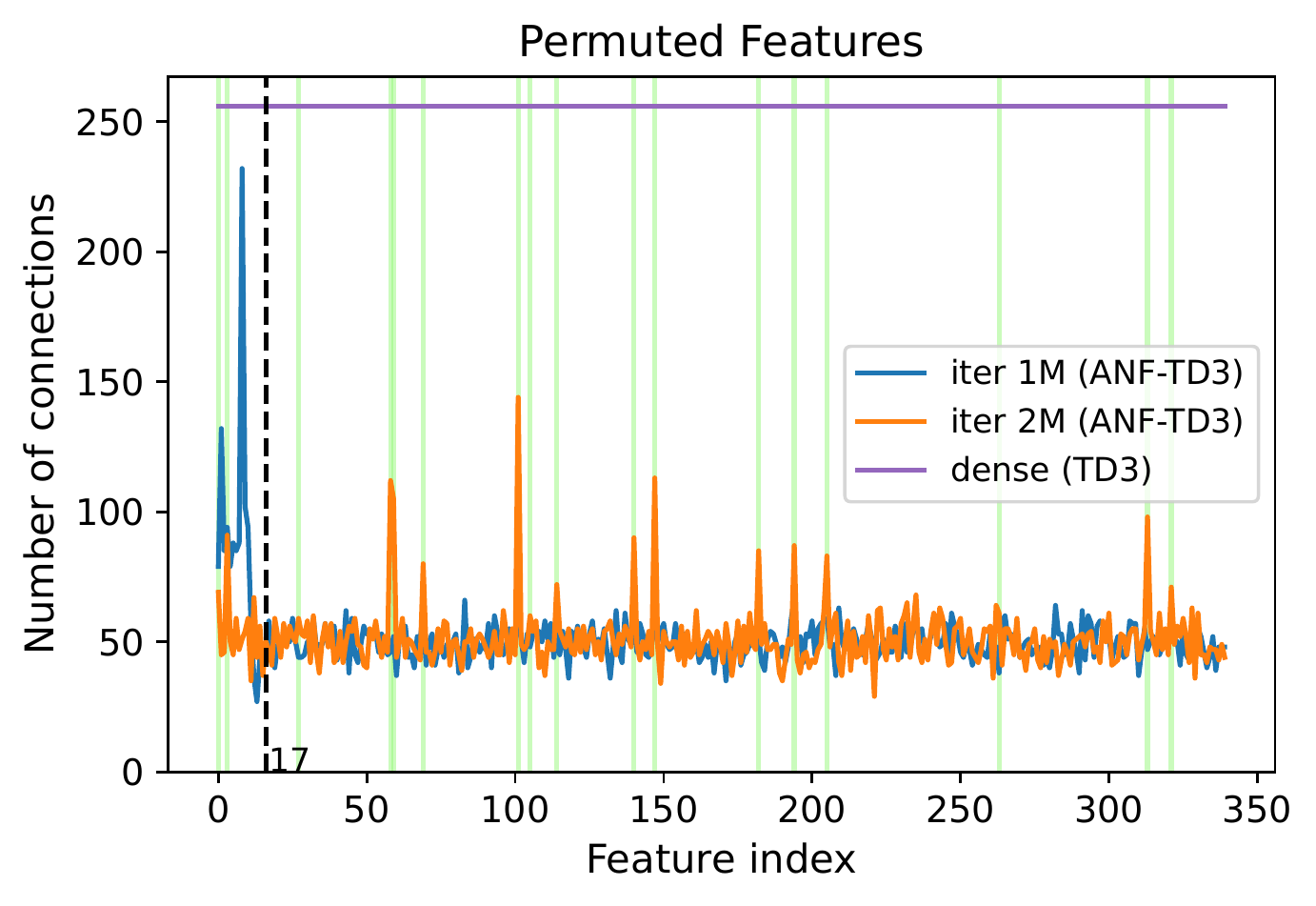}
    \caption{Number of connections per input neuron (feature) of the input layer of ANF-TD3's actor network, on HalfCheetah-v3 with 95\% noise features. After 1M iterations the agent has been trained on the initial setup (all 17 relevant features have index 0-16). The environment now changes and shuffles all input features with a fixed permutation. After 2M iterations the agent has adjusted to this new environment, and is able to find the new locations of the relevant features (shown by the green vertical lines).}
    \label{fig:permuted_actor_HalfCheetah}
\end{figure}

We see in Figure~\ref{fig:permuted_actor_HalfCheetah} that during training on the second sub-environment ANF drops the connections to input neurons that previously had task-relevant features (the leftmost features with index 0-16). It grows new connections to input neurons that currently have task-relevant features, represented by the green vertical lines.

\vspace{0.5cm}
\subsection{Louder noise graphs}
\label{sec:app-louder}

\begin{figure}[H]
    \centering
    \captionsetup{width=0.6\textwidth}
    \includegraphics[width=0.5\textwidth]{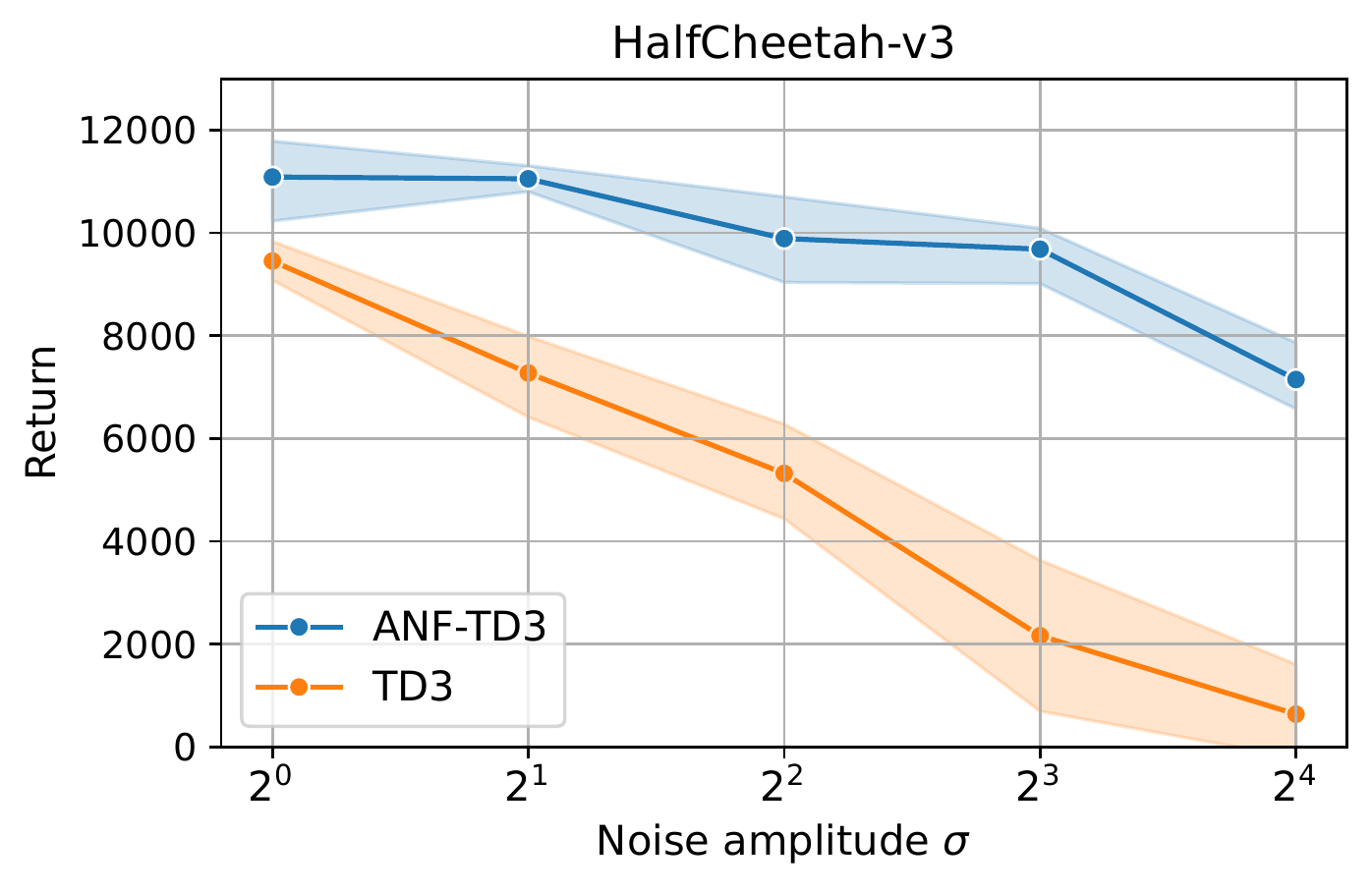}
    \vspace{-0.4cm}
    \caption{Performance of ANF and its baseline on ENEs with louder noise. 
    The noise features are sampled i.i.d. from $\mathcal{N}(0, \sigma^2)$. 
    Noise amplitude is increased exponentially, notice the log-scale on the horizontal axis. 
    This experiment uses 90\% noise features.
    A similar graph for ANF-SAC is shown in Figure~\ref{fig:louder1_16_SAC} of Section~\ref{sec:louder}.}
    \label{fig:louder1_16_TD3}
\end{figure}

Additional results on the louder noise experiments for TD3 are shown in Figure~\ref{fig:louder1_16_TD3}. 

\newpage
\subsection{Imitating real noise graphs}
\label{sec:app-imitate}

\begin{figure}[H]
    \centering
    \captionsetup{width=0.6\textwidth}
    \includegraphics[width=0.5\textwidth]{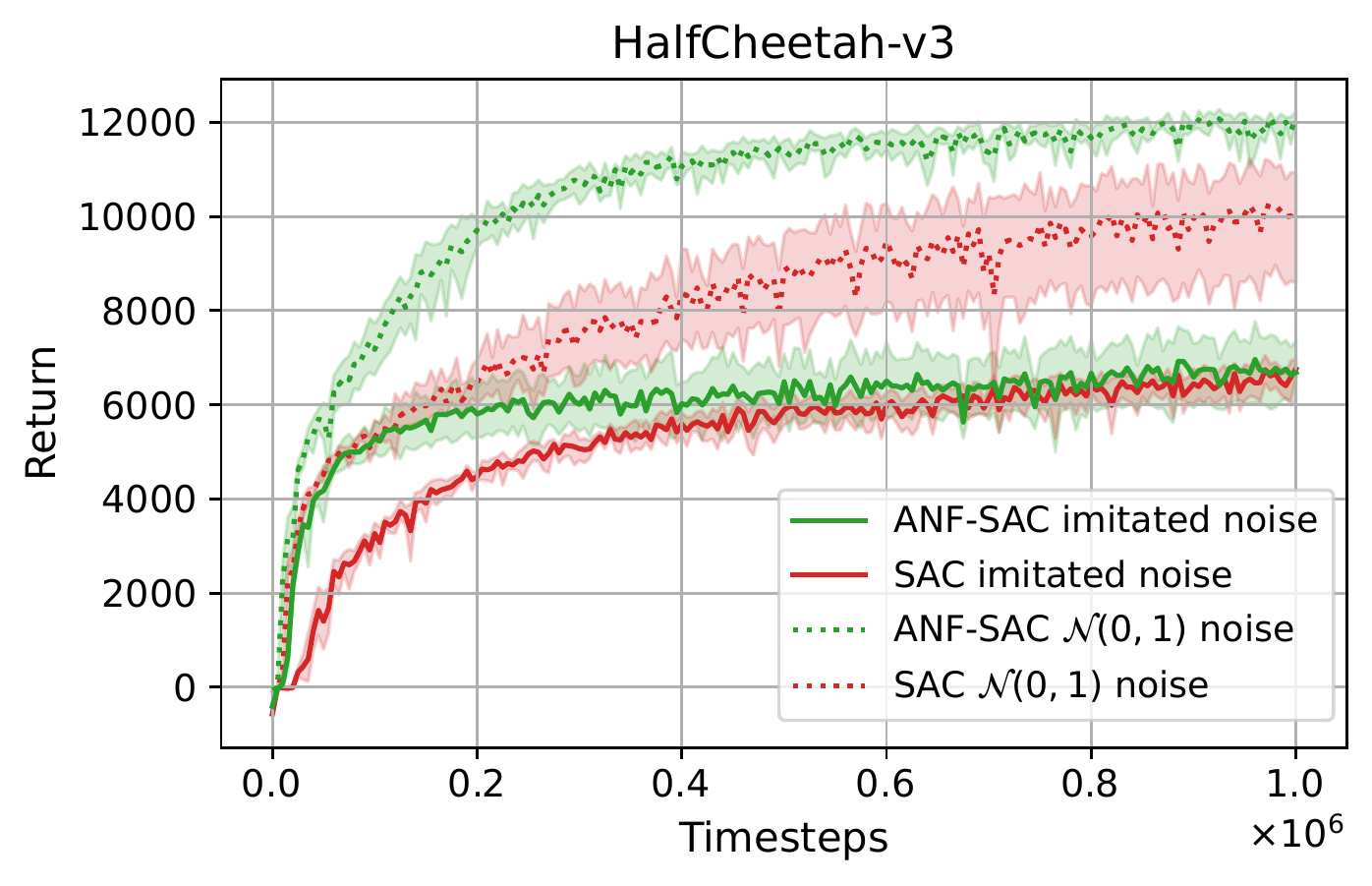}
    \vspace{-0.4cm}
    \caption{Learning curves for ANF-SAC and its baseline on the challenging ENE, where the noise features imitate the task-relevant features. This increases the difficulty, but ANF-SAC still learns much quicker than SAC. A similar graphs for ANF-TD3 is shown in Figure~\ref{fig:imitated_HalfCheetah-curves-TD3} of Section~\ref{sec:imitate}.}
    \label{fig:imitated_HalfCheetah-curves-SAC}
\end{figure}

The graph for ANF-SAC in the imitated noise experiment is shown in Figure~\ref{fig:imitated_HalfCheetah-curves-SAC}.

\vspace{1cm}
\subsection{Static ablation graphs}
\label{sec:app-static}

\begin{figure}[H]
    \centering
    \captionsetup{width=0.6\textwidth}
    \includegraphics[width=0.5\textwidth]{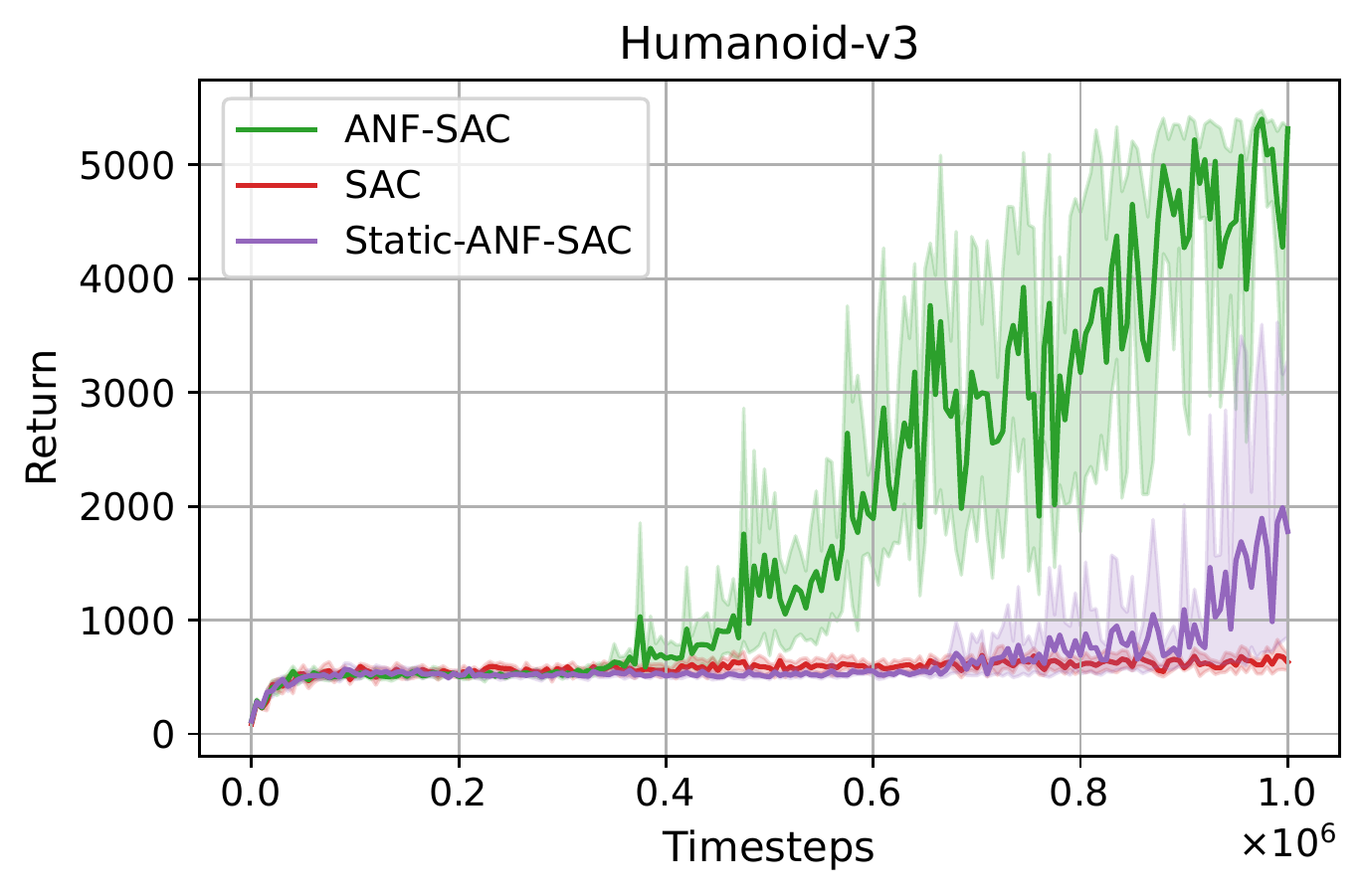}
    \vspace{-0.4cm}
    \caption{Comparison of ANF, which dynamically updates the sparse network topology, to its static sparse counterpart and the standard (fully dense) SAC. This experiment is run with 90\% noise features. A similar graph for TD3 is shown in Figure~\ref{fig:static_curves_fake90_Humanoid_TD3} of Section~\ref{sec:static}.}
    \label{fig:static_curves_fake90_Humanoid_SAC}
\end{figure}

Graphs for the static ablation study are shown in Figure~\ref{fig:static_curves_fake90_Humanoid_SAC}.

\newpage
\subsection{Sparsifying further}
\label{sec:app-sparser}

\begin{figure}[H]
    \centering
    \captionsetup{width=0.8\textwidth}
    \includegraphics[width=0.7\linewidth]{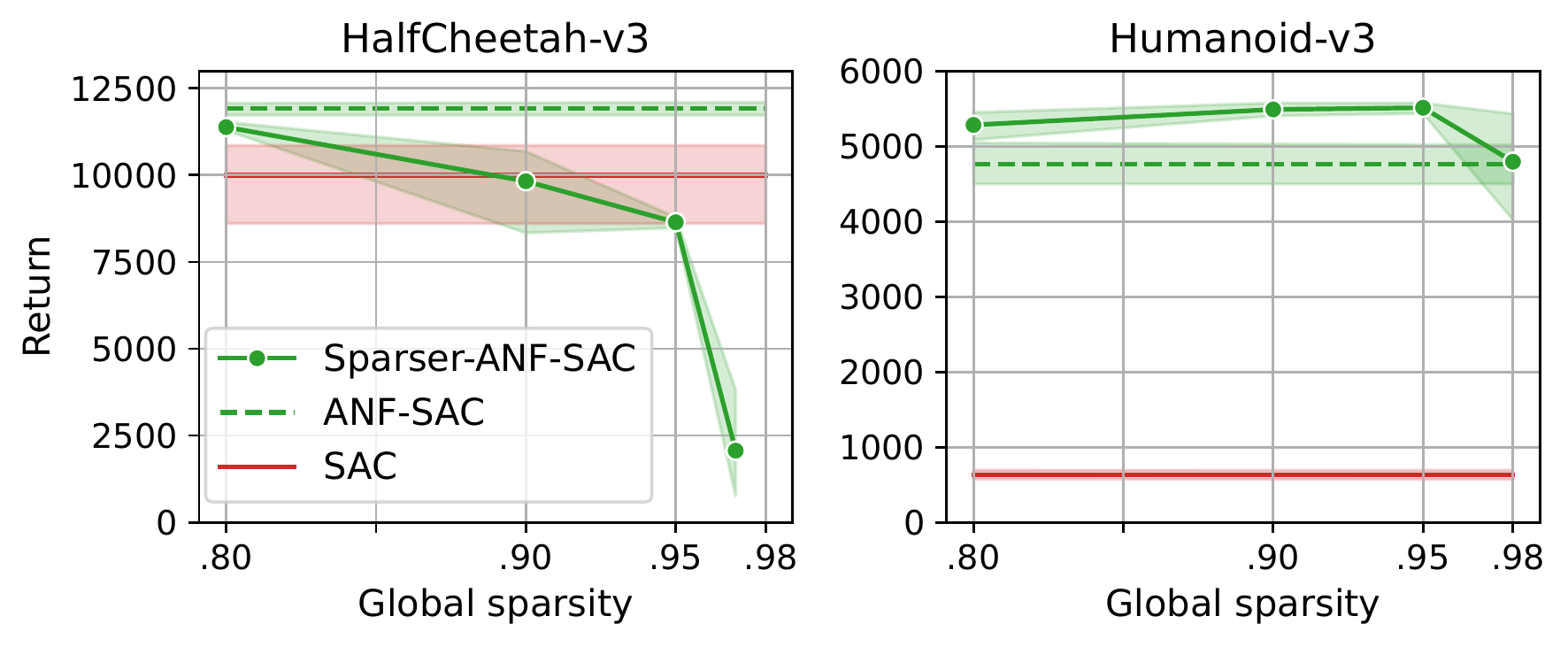}
    \vspace{-0.2cm}
    \caption{Performance of ANF-SAC and its sparser versions, on 90\% noise features. Sparser-ANF also prunes weights in the hidden layer, instead of only the input layer. Further sparsifying the network surprisingly improves performance for Humanoid (up to a certain point), but not for HalfCheetah. In both cases, it drastically reduces the size of the network. The graphs for ANF-TD3 are shown in Figure~\ref{fig:sparse_2envs_TD3} of Section~\ref{sec:sparse}.}
    \label{fig:sparse_2envs_SAC}
\end{figure}

Additional results for the experiments where we sparsified the agents even further are shown in Table~\ref{tab:overview_ext}, which is an extension of Table~\ref{tab:overview}, and in Figure~\ref{fig:sparse_2envs_SAC}. Notice that for HalfCheetah we could only go up to 97\% global sparsity with ANF-SAC. This is because 98\% turned out to be impossible for SAC on HalfCheetah with noise fraction $n_f=0.9$. The actor network of SAC has two output heads (in contrast to TD3, which only has one). This means that SAC's actor output layer has more connections (more than 2\% of the total possible number of connections). We keep the output layer dense in all these experiments, meaning that for SAC on HalfCheetah, 97\% global sparsity was the highest we could go.

\begin{table}[H]
\captionsetup{width=0.8\textwidth}
\caption{Comparison of different algorithms, along with their parameter counts for the actor network. The critic networks have comparable numbers of parameters.}
\label{tab:overview_ext}
\vspace{-0.2cm}
\setlength{\tabcolsep}{4pt}
\begin{tabular}{llrr}\toprule
Algorithm             & Environment     & Return ($\uparrow$) & \# Params. ($\downarrow$) \\ 
\midrule
ANF-TD3               & Walker2d-v3     & \textbf{4622.0}   & 75,776     \\ 
TD3                   & Walker2d-v3     & 3554.8             & 110,592   \\ 
Sparser(80\%)-ANF-TD3 & Walker2d-v3     & 4060.7            & 22,118     \\ 
Sparser(95\%)-ANF-TD3 & Walker2d-v3     & 2677.4            & \textbf{5,529}      \\ 
ANF-TD3               & Hopper-v3  & \textbf{3329.7}   & 71,936      \\ 
TD3                   & Hopper-v3  & 2941.8            & 94,464     \\ 
Sparser(80\%)-ANF-TD3 & Hopper-v3  & 3309.1            & 18,892      \\ 
Sparser(95\%)-ANF-TD3 & Hopper-v3  & 2807.2            & \textbf{4,723}       \\ 
\bottomrule         
\end{tabular}%
\end{table}

\subsection{Non-zero centered noise}

One of our anonymous reviewers pointed out that the zero-centered $\mathcal{N}(0, 1)$ noise simplifies ANF’s process of pruning connections to noise-features. We ran some extra experiments with non-zero-centered Gaussian noise to show empirically that: (i) non-zero-centered noise is indeed more challenging and (ii) ANF is still able to handle it better than dense networks.
The experiment is run on HalfCheetah-v3 with noise distribution $\mathcal{N}(\mu, 1)$, noise fraction $n_f=.98$, and 5 random seeds.

\begin{table}[H]
    \centering
    \captionsetup{width=0.8\textwidth}
    \caption{Experiment with non-zero-centered noise $\mathcal{N}(\mu, 1)$.}
    \label{tab:non-zero}
    \begin{tabular}{rcccc}
    \toprule
    Avg. returns     & $\mu=0$ & $\mu=1$ & $\mu=-2$ & $\mu=4$ \\
    \midrule
    ANF-SAC & 9250.4  &  5642.2  &  5047.8  &  636.9 \\
    SAC    &      6124.3 &   3744.4   &  419.1 &   $-$35.3 \\
    \bottomrule
    \end{tabular}
\end{table}


\paragraph{Theoretical perspective.} 
Even with non-zero-mean noise, we think the weights connected to noise-features will stay close to zero (and thus get pruned by ANF). Note that initial weights are small, and ANF’s newly grown weights start at 0. Since a noise-feature is irrelevant, its connections will receive mixed signals (positive gradient for some $<$state, action, reward$>$ tuples, negative gradient for others). This means it barely gets a chance to grow a large magnitude. In a simplified setting, we can prove a stronger claim;

\vspace{1em}
\textbf{Conjecture}: Weights connected to noise-features converge to 0 with gradient-descent, even for non-zero-centered noise. 

\begin{proof}
(Not a full proof, only for a simple setting.) We assume to be in the local neighborhood of the optimum. Suppose we have a function approximator $f(x_1,x_2) = w_1\cdot x_1 + w_2 \cdot x_2 = y_{\text{approx}}$, which is trying to estimate the true function $g(x_1)=a\cdot x_1=y$. Note: $g(\cdot)$ does not depend on $x_2$ (noise-feature).
\vspace{1em}

\noindent
Suppose we use mean-squared-error loss: $L = \frac12 (f(x_1,x_2) - y)^2$. Then the gradient (partial derivative) of weight $w_2$ is:
\begin{equation*}
\frac{ \textrm{d}L }{ \textrm{d}w_2} = ( f(x_1,x_2) - y ) \cdot x_2 = (w_1\cdot x_1 + w_2\cdot x_2 - a\cdot x_1)\cdot x_2. 
\end{equation*}
Assuming we’re near the optimum, i.e., $w_1 - a 
\approx 0$, we can rewrite $\frac{ \textrm{d}L }{ \textrm{d}w_2} \approx w_2 \cdot (x_2)^2$.
\vspace{1em}

\noindent
For any noise $x_2$: if $w_2$ is negative, its gradient is negative, and vice versa. We minimize MSE-loss, so gradient-descent moves in the direction: $-$gradient. Thus, $w_2$ will be updated toward zero.
\end{proof}

\vspace{4em}
\subsection{Matching the input-layer-sparsity-level with the noise-fraction}

From the problem setting of extremely noisy environments (ENEs) it seems beneficial to match the algorithm's input-layer-sparsity-level with the noise-fraction of the ENE. We ran this experiment but omitted it from the main body of the paper for the following reasons:
\begin{enumerate}[(i)]
    \item performance is similar (see Table~\ref{tab:matching} below, it’s challenging to prune all connections to noise-features, so it may be useful to have some surplus of connections for task-relevant features),
    \item we did not want to assume that the agent knows the noise-fraction. (The input-layer-sparsity-level could be an adaptive parameter, which is mentioned as potential future work in the paper.)
\end{enumerate}
Instead, we kept the input-layer-sparsity at a well-working 80\% to have a generally applicable algorithm. 

\begin{table}[H]
    \centering
    \captionsetup{width=0.65\textwidth}
    \caption{Experiments matching input-layer-sparsity-levels with noise-fractions (policy = ANF-SAC, env = HalfCheetah-v3, num\_seeds = 5).}
    \label{tab:matching}
    \begin{tabular}{rcccc}
    \toprule
    Avg. returns     & $n_f=.90$ &  $n_f=.95$ & $n_f=.98$ & $n_f=.99$ \\
    \midrule
    matching input-layer-sparsity & 11041.1  &  10259.0  &  10090.3  &   8641.2   \\
    80\% input-layer-sparsity   &    11913.8  &   9920.5  &   9250.4 &   10007.2  \\
    \bottomrule
    \end{tabular}
\end{table}


\newpage
\section{Distributions of original features}
\label{sec:app-distr}

\begin{figure}[H]
    \centering
    \captionsetup{width=0.95\textwidth}
    \includegraphics[width=0.9\textwidth]{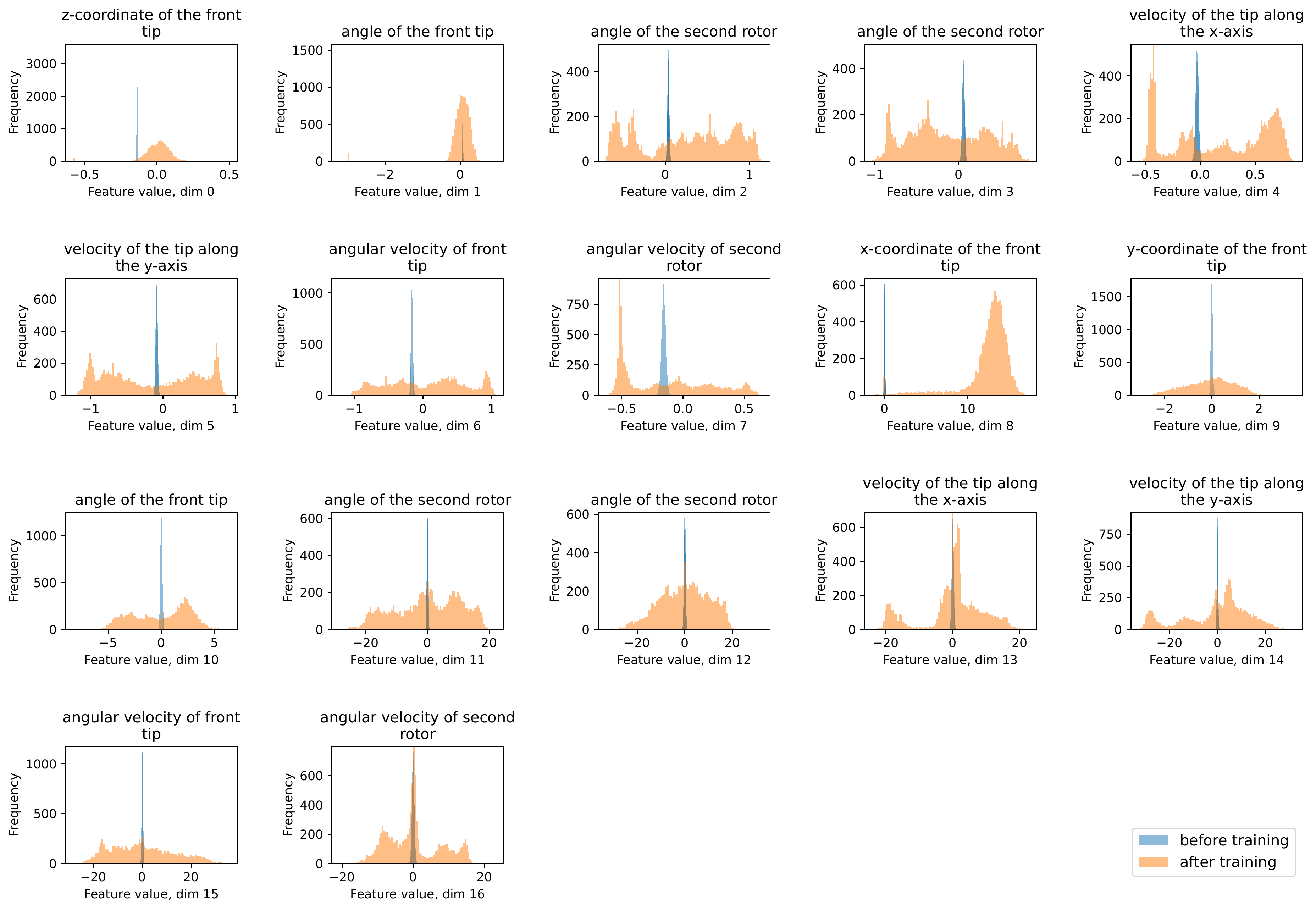}
    \caption{Distributions of the 17 original features of HalfCheetah-v3, after running 10K steps with a trained and an untrained model (ANF-TD3 on 90\% noise features). Note that the distributions tend to widen significantly after training.}
    \label{fig:feats_distribution_HalfCheetah_all_dims}
\end{figure}

\begin{figure}[H]
    \centering
    \captionsetup{width=0.95\textwidth}
    \includegraphics[width=0.9\textwidth]{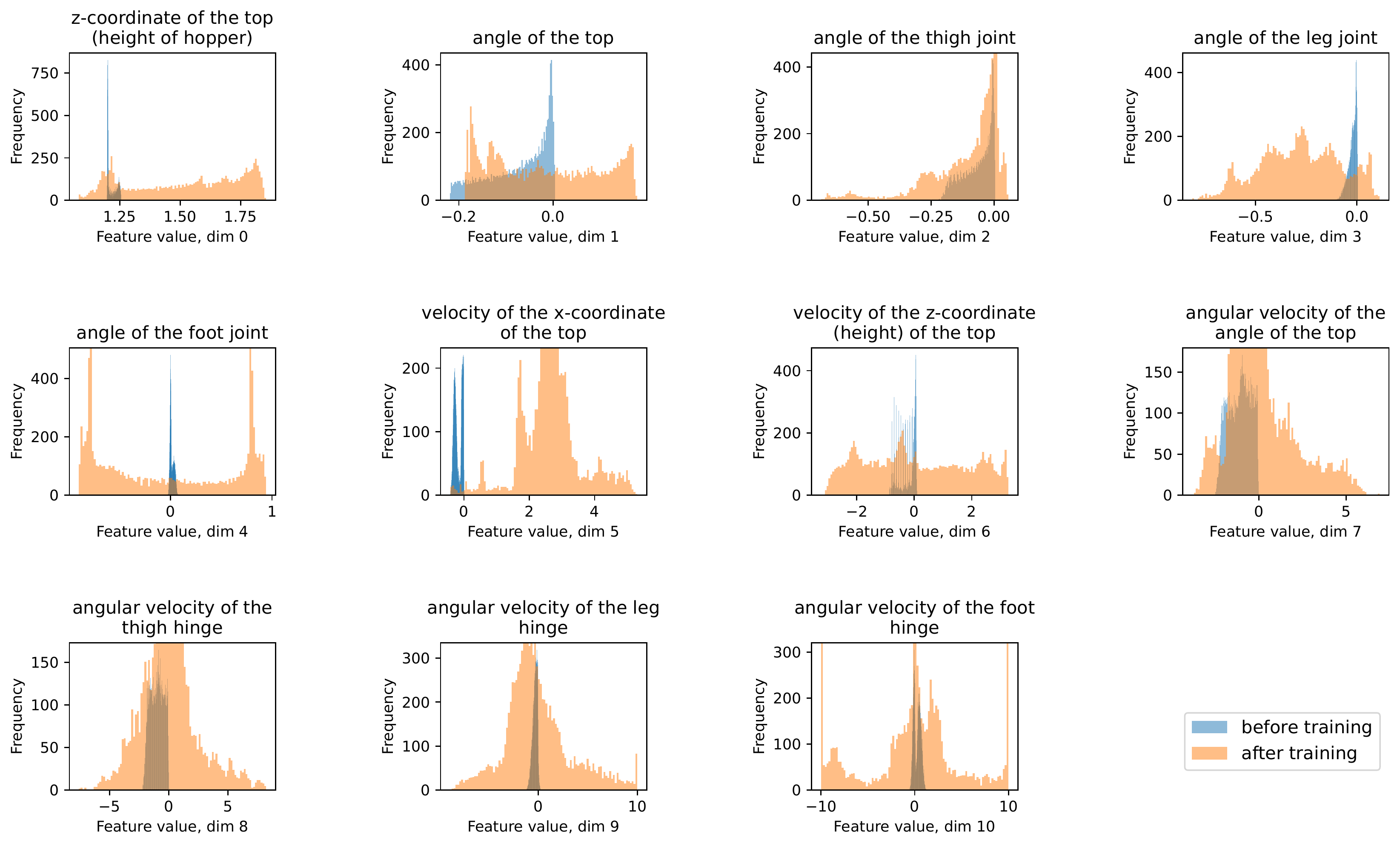}
    \caption{Distribution of the 11 original features of Hopper-v3, after running 10K steps with a trained and an untrained model (ANF-TD3 on 90\% noise features).}
    \label{fig:feats_distribution_Hopper_all_dims}
\end{figure}

\begin{figure}[H]
    \centering
    \captionsetup{width=0.95\textwidth}
    \includegraphics[width=0.9\textwidth]{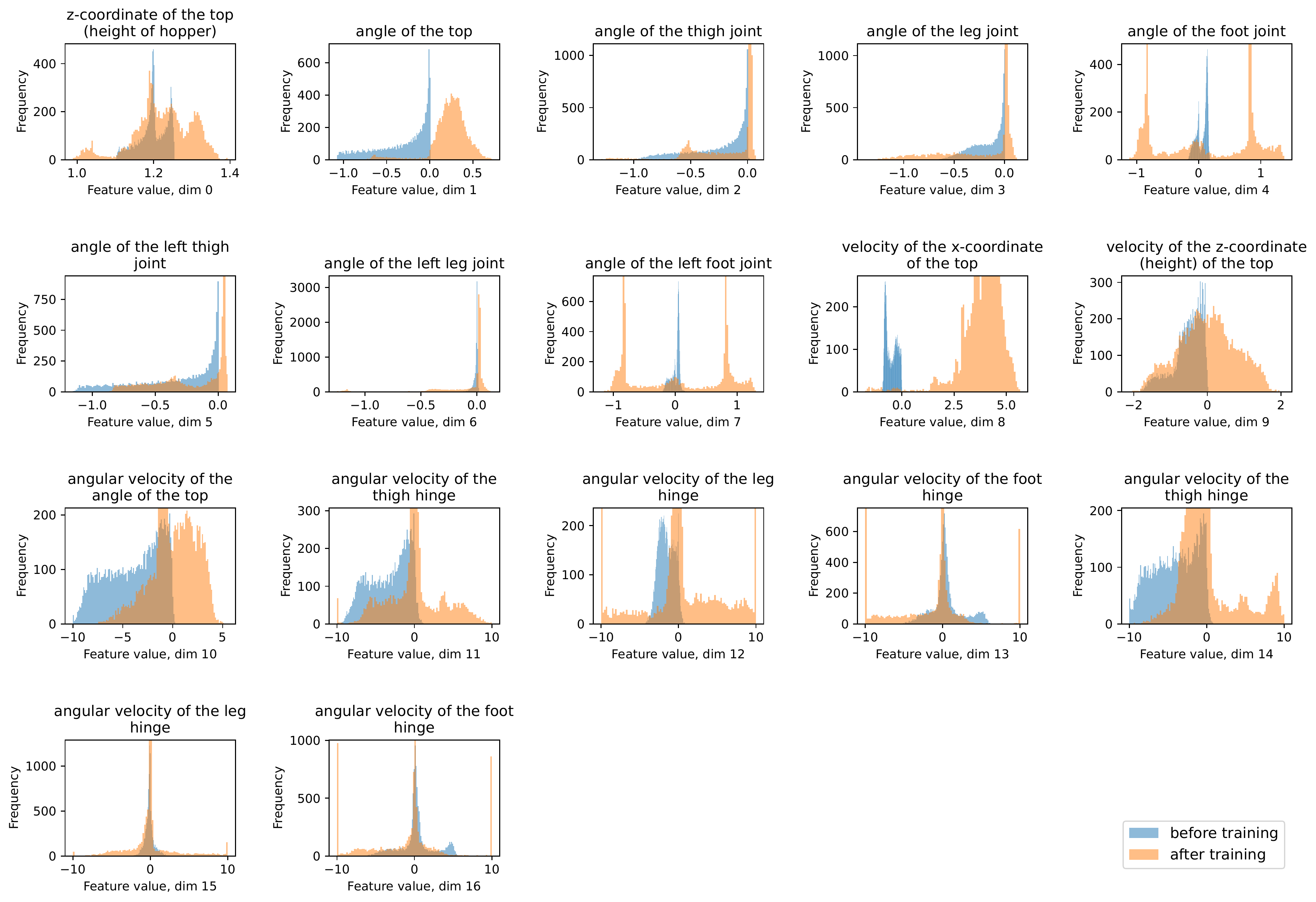}
    \caption{Distribution of the 17 original features of Walker2d-v3, after running 10K steps with a trained and an untrained model (ANF-TD3 on 90\% noise features).}
    \label{fig:feats_distribution_Walker2d_all_dims}
\end{figure}

\begin{figure}[H]
    \centering
    \captionsetup{width=0.95\textwidth}
    \includegraphics[width=0.9\textwidth]{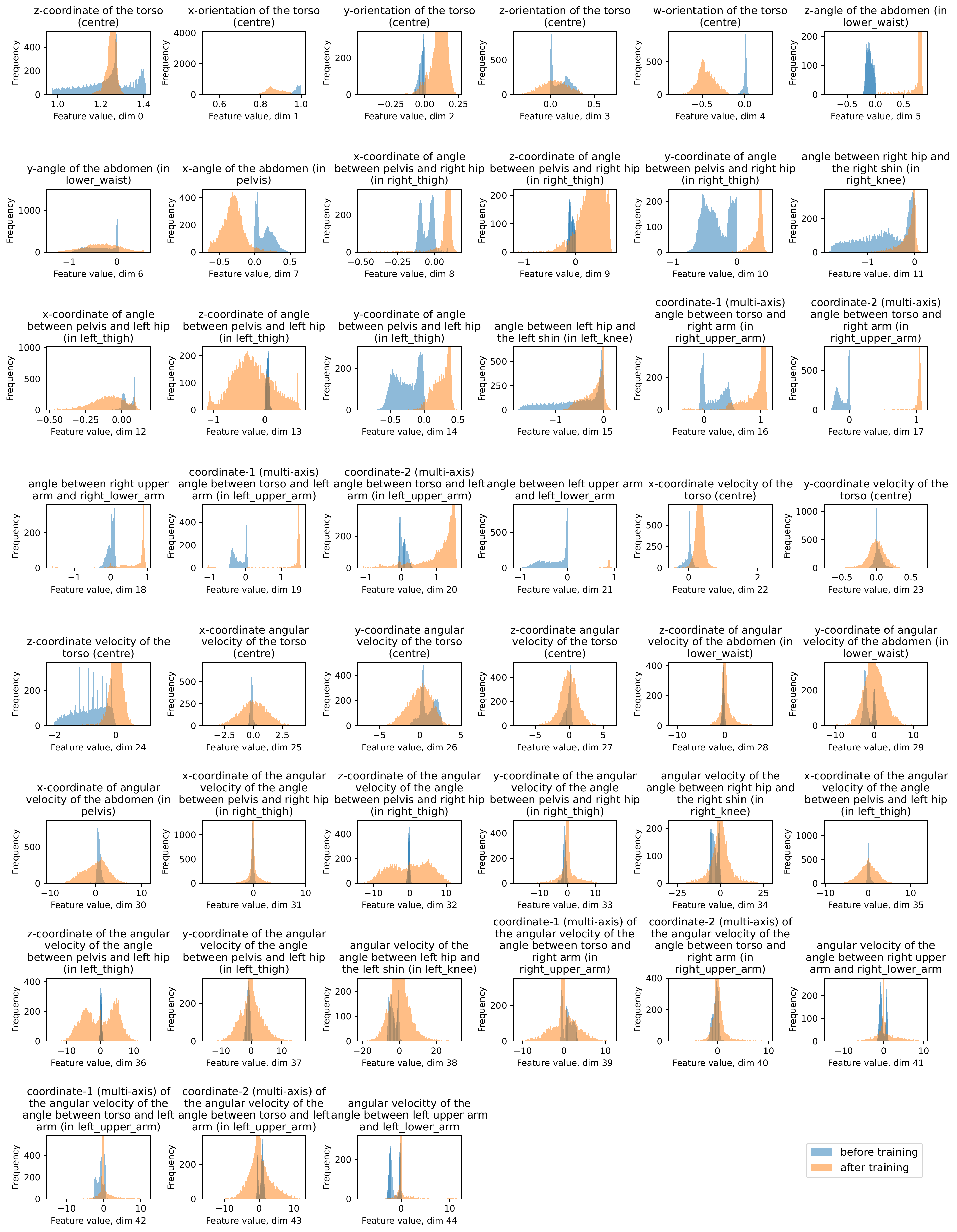}
    \caption{Distribution of the first 45 original features of Humanoid-v3. A figure showing histograms for all 376 dimensions is available for \href{https://www.dropbox.com/s/rcw9ihchq1oa5db/features_distribution_Humanoid_all.pdf?dl=0}{download online}.}
    \label{fig:feats_distribution_Humanoid_45_dims}
\end{figure}

Here we present the distributions of the original state features of the MuJoCo Gym environments. These histograms are generated by taking a trained or untrained agent and letting it run on an evaluation environment for $10,000$ steps while recording the feature values.
We used the orange (after-training) distributions to generate realistic noise features in the experiments of Section~\ref{sec:imitate}. Note that the distributions differ a lot depending on whether the agent is trained or not; the non-stationarity of the data distribution in RL is quite evident. See Figures~\ref{fig:feats_distribution_HalfCheetah_all_dims}, \ref{fig:feats_distribution_Hopper_all_dims}, \ref{fig:feats_distribution_Walker2d_all_dims}, \ref{fig:feats_distribution_Humanoid_45_dims}. The titles of the features are taken from the environment's documentation.\footnote{See \url{https://www.gymlibrary.dev/environments/mujoco/}.}

\end{document}